\documentclass[letterpaper]{article} 
\usepackage[preprint]{aaai2027}  
\usepackage[hyphens]{url}  
\usepackage{graphicx} 
\usepackage[most]{tcolorbox}
\usepackage{array}
\usepackage{natbib}  
\usepackage{caption} 
\usepackage{algorithm}
\usepackage{algorithmic}

\usepackage{newfloat}
\usepackage{listings}
\DeclareCaptionStyle{ruled}{labelfont=normalfont,labelsep=colon,strut=off} 
\floatstyle{ruled}
\newfloat{listing}{tb}{lst}{}
\floatname{listing}{Listing}

\usepackage{booktabs}
\usepackage[table]{xcolor}

\newcommand{\rwyes}{\textcolor{green!50!black}{\boldmath$\surd$}}
\newcommand{\rwno}{\textcolor{red!70!black}{\boldmath$\times$}}

\usepackage{subcaption}

\usepackage{amsmath}

\newenvironment{promptbox}[1]{%
\begin{tcolorbox}[
  enhanced,
  breakable,
  colback=white,
  colframe=black!60,
  colbacktitle=black!60,
  coltitle=white,
  title={#1},
  fonttitle=\bfseries,
  toptitle=1mm,
  bottomtitle=1mm,
  lefttitle=2mm,
  righttitle=2mm,
  boxrule=0.7pt,
  arc=2mm,
  outer arc=2mm,
  left=1.5mm,
  right=1.5mm,
  top=2mm,
  bottom=2mm,
  boxsep=0pt,
  width=\linewidth
]
\fontsize{9pt}{9.2pt}\selectfont
}{%
\end{tcolorbox}
}

\title{CoCoBench: A Cooperative Coordination Benchmark for Embodied Multi-Agent Task Planning}
\author{
    Yang Chen\textsuperscript{\rm 1,2},
    Ye-Xin Xie\textsuperscript{\rm 1},
    Lirong Che\textsuperscript{\rm 2,3},
    Danyang Peng\textsuperscript{\rm 1},
    Yuzhe Yang\textsuperscript{\rm 2},
    Peiwen Lin\textsuperscript{\rm 2},\\
    Xu Cao\textsuperscript{\rm 2},
    Chuang Wang\textsuperscript{\rm 2},
    Lei Yuan\textsuperscript{\rm 1},
    Jian Su\textsuperscript{\rm 2}\corresponding,
    Lan-Zhe Guo\textsuperscript{\rm 1}\corresponding
}
\affiliations{
    \textsuperscript{\rm 1}Nanjing University\\
    \textsuperscript{\rm 2}AgiBot\\
    \textsuperscript{\rm 3}Tsinghua University
}

\begin{document}

\maketitle

\begin{abstract}
Agent systems powered by multimodal large language models (MLLMs) have advanced rapidly in recent years, yet existing embodied-agent benchmarks still lack fine-grained diagnostics for multi-agent coordination. Most benchmarks either focus on single-agent task completion or summarize multi-agent behavior with overall task success rates, which can obscure coordination failures such as duplicated work, violations of ordering constraints, resource contention, and desynchronized handoffs. In this paper, we introduce \textbf{CoCoBench}, a construct-level benchmark for evaluating multi-agent embodied coordination in executable household tasks. CoCoBench contains 897 oracle-validated instances organized around four recurring coordination constructs: task allocation, sequential ordering, mutual exclusion, and handoff coordination. In addition to task success rate, CoCoBench provides construct-level scores that measure whether agents coordinate effectively. We evaluate 11 leading MLLMs across different coordination modes, observation inputs, and numbers of agents. The results show that coordination ability is highly construct-specific: strong overall performance does not imply balanced competence across different coordination types. These findings point to new directions for designing targeted model architectures and improving multi-agent coordination ability. The code is available at \url{https://github.com/AgibotGeneral/CoCoBench}.
\end{abstract}


\begin{figure}[t!]
\centering
\includegraphics[width=\linewidth]{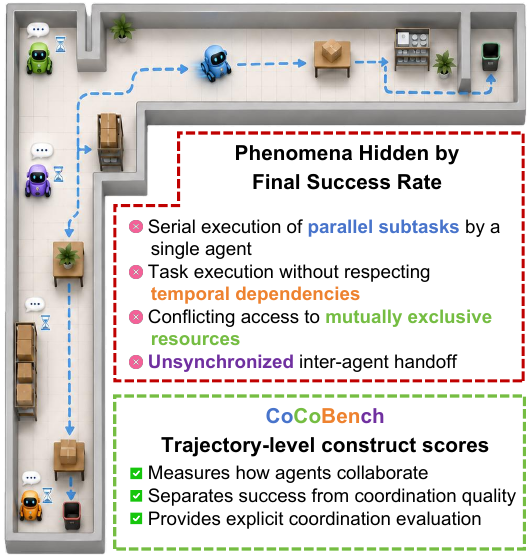}
\caption{\textbf{Motivation of CoCoBench.} Existing benchmarks either focus on single-agent task success or multi-agent final completion, whereas CoCoBench explicitly evaluates coordination ability through construct-level metrics.}
\label{fig:intro}
\end{figure}

\begin{figure*}[t]
\centering

\begin{minipage}[c]{0.40\textwidth}
    \centering
    \includegraphics[
        width=\linewidth,
        trim=22pt 24pt 24pt 18pt,
        clip
    ]{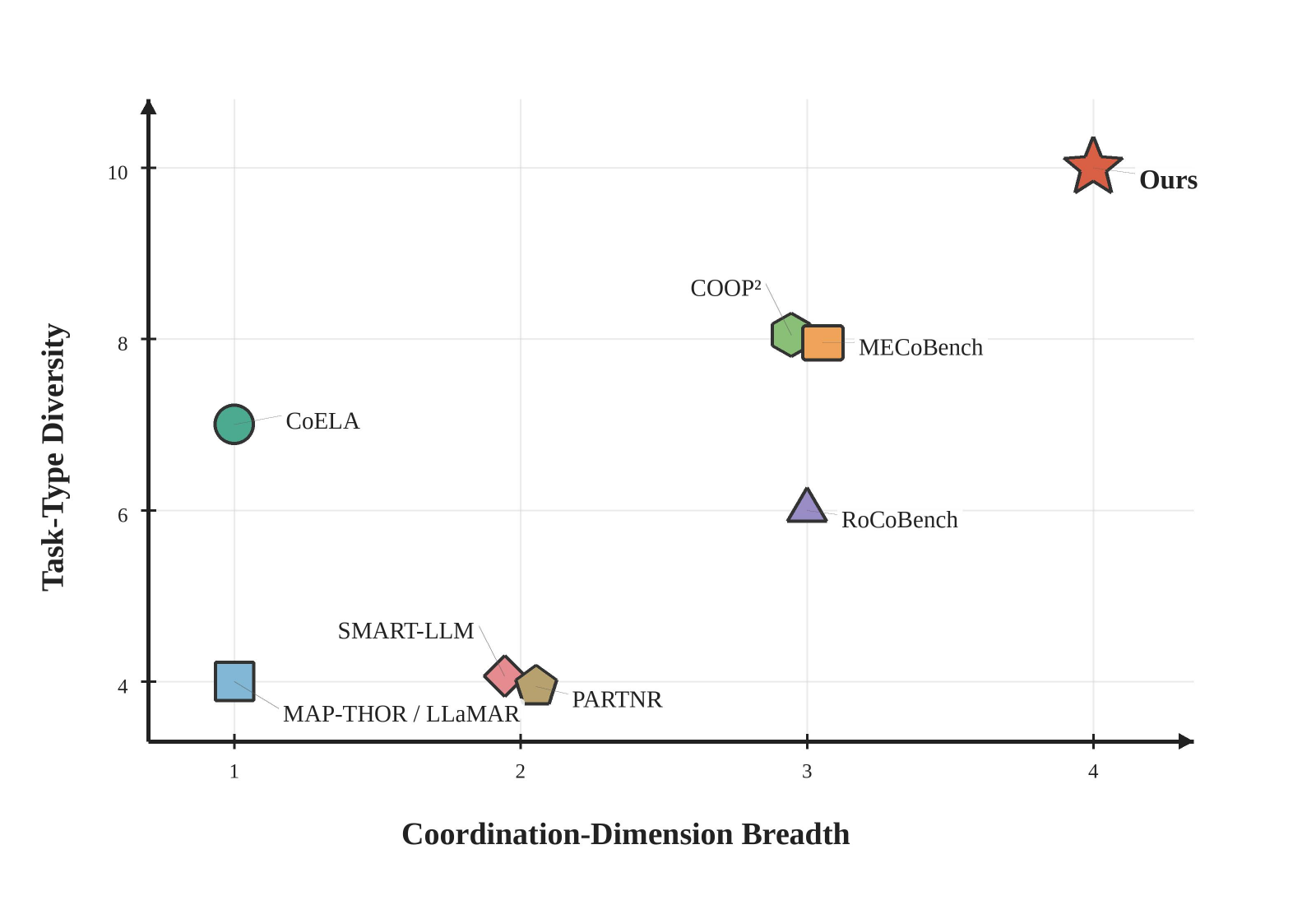}
\end{minipage}
\hfill
\begin{minipage}[c]{0.56\textwidth}
    \centering
    \scriptsize
    \setlength{\tabcolsep}{3pt}
    \renewcommand{\arraystretch}{1.16}

    \resizebox{\linewidth}{!}{%
    \begin{tabular}{lccccccc}
        \toprule
        \textbf{Benchmark} & \textbf{Dims.} & \textbf{Cases}
        & \textbf{Cent.} & \textbf{Dec.} & \textbf{MM}
        & \textbf{Role} & \textbf{TS} \\
        \midrule

        CoELA~\cite{coela}
        & T & 44
        & \rwno & \rwyes & \rwno & \rwno & \rwno \\

        RoCoBench~\cite{roco}
        & T/S/H & 6
        & \rwno & \rwyes & \rwno & \rwyes & \rwno \\

        SMART-LLM~\cite{smartllm}
        & T/S & 36
        & \rwyes & \rwno & \rwno & \rwyes & \rwno \\

        MAP-THOR~\cite{llamar}
        & T & 225
        & \rwyes & \rwno & \rwyes & \rwno & \rwno \\

        PARTNR~\cite{partnr}
        & T/S & 1000
        & \rwyes & \rwyes & \rwno & \rwno & \rwno \\

        COOP$^2$~\cite{coop2}
        & T/S/H & 36
        & \rwyes & \rwyes & \rwno & \rwyes & \rwno \\

        MECoBench~\cite{mecobench}
        & T/S/H & 192
        & \rwyes & \rwyes & \rwyes & \rwno & \rwno \\

        \midrule
        \textbf{CoCoBench (Ours)}
        & \textbf{T/S/M/H} & \textbf{897}
        & \rwyes & \rwyes & \rwyes & \rwyes & \rwyes \\

        \bottomrule
    \end{tabular}%
    }
\end{minipage}

\caption{\textbf{Comparison with related embodied benchmarks.}
\textbf{Left:} coordination-dimension breadth versus task-type diversity,
where breadth counts the coverage of T, S, M, and H.
\textbf{Right:} benchmark attributes, where \textbf{Cases} denotes the
number of evaluation cases; \textbf{Dims.} lists the covered coordination
dimensions; \textbf{Cent.}/\textbf{Dec.}, \textbf{MM}, and \textbf{Role}
indicate centralized/decentralized modes, multimodal observations, and
heterogeneous roles, respectively. \textbf{TS} denotes trajectory-level
construct scores. T/S/M/H represent task allocation, sequential ordering,
mutual exclusion, and handoff coordination.}
\label{fig:related-work}
\end{figure*}

\section{Introduction}
\label{sec:introduction}

Multimodal large language models (MLLMs) are increasingly used as the brains of embodied agents, enabling high-level planning, grounded reasoning, and embodied control~\citep{codeaspolicies,embodiedgpt,embodiedagent,embodiedllm,generalistagents}. Meanwhile, embodied agents are moving beyond isolated task execution toward collaborative teams operating in shared physical environments, in line with broader progress in multi-agent systems and LLM-based collaboration~\citep{metagpt,scalingmultiagent}. In household scenarios, many tasks are naturally collaborative: agents with different embodiments and capabilities may need to distribute objects across containers, follow temporal dependencies around a workstation, coordinate access to shared tools, or transfer objects through intermediate handoff points. These tasks require more than object recognition and primitive skill execution. Even with correct perception and valid low-level skills, a team may fail by duplicating work, closing a container too early, competing for the same tool, or leaving a handoff partner idle.

Existing embodied-agent benchmarks provide rich evaluations of task and motion planning~\citep{vlabench,embodiedbench}, and recent work has begun to study multi-agent embodied tasks and collaborative agent behavior~\citep{cordialsync,collabovercooked}. However, most evaluations still rely on unified success rates or efficiency metrics. Such aggregate metrics make multi-agent behavior difficult to diagnose: failures may arise from poor allocation, ordering violations, resource contention, handoff desynchronization, or low-level execution errors, while successful final states may still hide repeated invalid attempts, unnecessary waiting, or avoidable conflicts. Therefore, embodied collaboration should be evaluated not only by whether the goal is achieved, but also by how agents coordinate to achieve it, as illustrated in Figure~\ref{fig:intro}.

In this work, we introduce \textbf{CoCoBench}, a construct-level benchmark with 897 instances for evaluating multi-agent embodied coordination. CoCoBench is organized around four recurring sources of coupling in collaborative household tasks: \emph{who} performs which subtasks, \emph{when} actions must follow temporal order, \emph{whether} agents can safely share resources, and \emph{how} intermediate objects are transferred between roles.

CoCoBench has two key designs. First, it instantiates four coordination constructs: \textbf{task allocation}, where agents divide independent subtasks; \textbf{sequential ordering}, where agents respect causal preconditions; \textbf{mutual exclusion}, where agents avoid conflicts over exclusive shared resources; and \textbf{handoff coordination}, where agents cooperate through an intermediate buffer in a producer-consumer pattern. Beyond task success rate, CoCoBench reports construct-level scores that measure allocation efficiency, precondition compliance, conflict-free resource scheduling, and handoff rhythm, thereby distinguishing effective coordination from merely reaching the final state. Second, all tasks use a configurable high-level skill interface covering household interactions such as navigation, pickup, placement, and slicing. Rather than defining tasks only by object categories or room layouts, CoCoBench defines them by coordination constructs that induce specific collaboration demands, while keeping every instance grounded in executable household activities.

We evaluate 11 leading MLLMs on CoCoBench. The results show that multi-agent embodied coordination is highly structured: strong aggregate performance does not imply balanced competence across coordination constructs; centralized state aggregation remains crucial; and simple communication only partially narrows the gap to centralized planning. Under CoCoBench's high-level skill interface, removing visual observations does not substantially degrade performance, suggesting that the main bottleneck lies in symbolic team-level planning rather than low-level visual recognition. Performance also decreases as team size grows, especially for weaker open-weight models. Diagnostic analyses further show that task success alone is insufficient, as successful episodes may still diverge from legal coordination.

Our contributions are threefold. \textbf{First}, we formalize multi-agent embodied coordination as trajectory-level joint planning and identify recurring sources of coordination demand. \textbf{Second}, we build CoCoBench, an 897-instance benchmark that instantiates four measurable coordination constructs in executable household tasks and provides corresponding construct-level scores. \textbf{Third}, we systematically evaluate MLLMs under different experimental settings, revealing coordination-specific failure modes hidden by aggregate success metrics.

\begin{figure*}[t!]
  \centering
  \includegraphics[width=\linewidth]{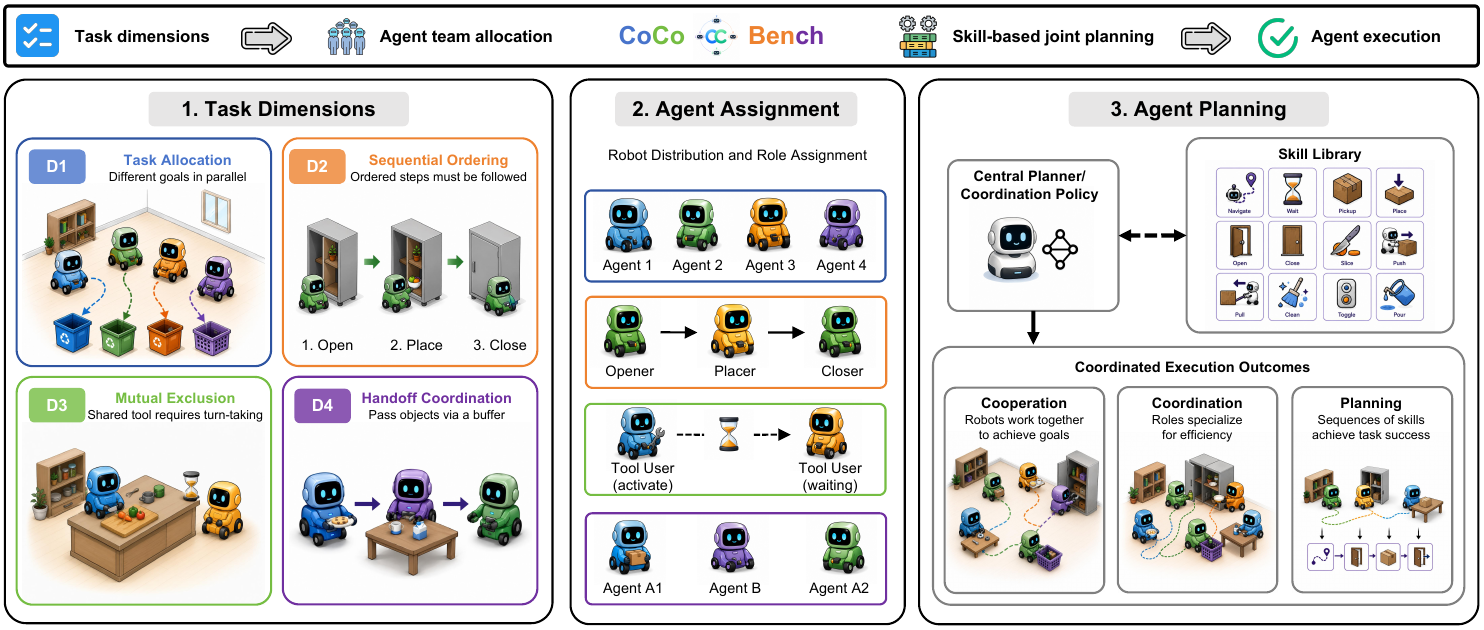}
  \caption{\textbf{Overview of the CoCoBench framework.} CoCoBench evaluates multi-agent embodied coordination through a three-stage pipeline: construct-level task design, agent assignment, and skill-based joint planning. The benchmark covers four coordination constructs and maps them to executable household tasks with explicit agent roles and high-level skill interfaces.}
  \label{fig:framework}
\end{figure*}

\section{Related Work}

\paragraph{Embodied Agents and Benchmarks.}
Recent agent research has increasingly used large language models and multimodal large language models as the brains of embodied agents, enabling task and motion planning~\citep{saycan,palme}, navigation~\citep{internvln}, and open-world game exploration~\citep{voyager,gamma-world}. Household environments are a major application domain for embodied agents. To explore the capability boundaries of models in planning tasks, prior work has developed rich simulation platforms~\citep{habitat,ai2thor} and evaluation benchmarks~\citep{alfred,alfworld,eai,embodiedbench}. Despite this progress, these benchmarks mainly evaluate single agents and provide limited support for task allocation, progress coordination, or joint execution among multiple agents.

\paragraph{Embodied Multi-Agent Benchmarks.}
Embodied multi-agent benchmarks extend evaluation to collaborative household tasks~\citep{coela,roco,smartllm,llamar,coop2,partnr}, robotic manipulation~\citep{robotwin,robotwin2}, and open-world environments~\citep{teamcraft,minecollab}. Among existing benchmarks, MECoBench~\citep{mecobench} is the work most closely related to ours. It studies multimodal embodied cooperation through parallel and sequential task structures, varying team sizes, coordination modes, and communication settings. Despite these advances, existing task taxonomies typically cover only a limited set of high-level coordination structures or conflate multiple dependency mechanisms within a single task, limiting mechanism-specific diagnosis. In contrast, CoCoBench decomposes task-level dependencies into explicit coordination mechanisms and provides construct-level scores for each mechanism, thereby distinguishing general task-execution failures from mechanism-specific coordination failures. Figure~\ref{fig:related-work} summarizes this comparison.

\section{Problem Formulation}
\label{sec:problem-formulation}

We formulate multi-agent embodied coordination as high-level joint decision making in a shared interactive environment.  A task instance is specified by
\begin{equation}
  x=(\mathcal{E},\mathcal{A},\mathcal{S},\mathcal{O},\mathcal{U},\Delta,\Phi,H,d),
\end{equation}
where $\mathcal{E}$ is a household scene, $\mathcal{A}=\{a_i\}_{i=1}^{N}$ is a team of $N$ embodied agents, $\mathcal{S}$ is the environment state space, $\mathcal{O}=\{O_i\}_{i=1}^{N}$ contains the agents' observation functions, $\mathcal{U}$ is the fixed set of executable high-level skills, $\Delta$ is the shared executor for low-level navigation and manipulation, $\Phi$ is a set of goal predicates, $H$ is the step budget, and $d$ denotes the target coordination construct.  At step $t$, agent $a_i$ observes $o_t^i=O_i(s_t)$.  A high-level coordination policy maps the agents' observation histories to a joint skill decision,
\begin{equation}
  \mathbf{u}_t=(u_t^1,\ldots,u_t^N)
  \sim \pi(\cdot\mid o_{\le t}^{1},\ldots,o_{\le t}^{N}),
  \qquad u_t^i\in\mathcal{U}.
\end{equation}
The executor applies these skills in the environment and records execution events, producing a trajectory
\begin{equation}
  \tau=(s_0,\mathbf{o}_0,\mathbf{u}_0,e_0,s_1,\ldots,s_H),
\end{equation}
where $e_t$ summarizes simulator-grounded events such as invalid skill use, failed preconditions, resource conflicts, redundant actions, and waiting.  The episode succeeds when the final state satisfies every goal predicate:
\begin{equation}
  \mathrm{Succ}_{x}(\tau)=\mathbf{1}\left[\forall \phi\in\Phi,\ \phi(s_H)=1\right].
\end{equation}

The key distinction from single-agent embodied planning is coordination coupling.  Let $F_x(u,s)$ denote the feasibility, goal contribution, or execution cost of skill $u$ in state $s$, and let $\Delta(s,u)$ denote the state reached after applying a skill through the executor.  A task is coordination-coupled if there exist two agents $a_i\ne a_j$ and skills $u_t^i,u_{t'}^j$ such that
\begin{equation}
  F_x(u_t^i,s_t)\ne F_x\!\left(u_t^i,\Delta(s_t,u_{t'}^j)\right).
\end{equation}
That is, one agent's action changes whether another agent's action is valid, useful, or efficient.  Such coupling arises from shared objects, spatial exclusion, temporal preconditions, exclusive resources, and intermediate states produced by one agent and consumed by another.

CoCoBench instantiates these couplings as construct-specific task families and evaluates the induced trajectory along two complementary axes:
\begin{equation}
  \mathrm{Eval}(\pi;x)=
  \left(\mathrm{Succ}_{x}(\tau),\ C_{d,x}(\tau)\right),
  \qquad \tau\sim\pi(\cdot\mid x).
\end{equation}
Here, $C_{d,x}(\tau)\in[0,1]$ is a construct-specific coordination score computed from the recorded execution events.  This formulation keeps final task completion separate from coordination quality, allowing CoCoBench to diagnose trajectories that reach the goal through illegal or inefficient behavior. The benchmark therefore targets team-level planning under a fixed skill interface rather than isolated primitive skill execution or a single blended reward.

\section{Benchmark Design}
\label{sec:benchmark-construction}

CoCoBench instantiates this formulation in AI2-THOR~\cite{ai2thor}, with each instance targeting one coordination construct and evaluated from recorded trajectories, as illustrated in Figure~\ref{fig:framework}.

\subsection{Coordination Constructs and Scores}
\label{sec:benchmark-composition}

CoCoBench targets four collaboration patterns that repeatedly arise in household work.  These constructs are defined by how agents are coupled through the environment rather than by a specific room or object category: agents may split independent chores, depend on a teammate's earlier action, compete for an exclusive tool, or pass objects through an intermediate station.  This abstraction lets us evaluate coordination as a capability while grounding every instance in concrete household activities.

In addition to task success, each trajectory receives a construct score \(C_d\in[0,1]\), where \(d\in\{\mathrm{D1},\mathrm{D2},\mathrm{D3},\mathrm{D4}\}\) denotes the target coordination construct. Success rate measures whether the final state satisfies the task goal, whereas \(C_d\) evaluates the quality of the construct-relevant coordination behavior observed along the trajectory, independently of task completion. Thus, a high \(C_d\) does not imply task success, and the two metrics should be interpreted jointly. Trajectories with \(n=0\), \(a=0\), or \(h=0\) are excluded due to insufficient signal.

\textbf{D1: Task allocation} covers chores that can be completed in parallel, such as sorting objects into target receptacles.  The coordination target is balanced allocation.  Let $M$ be the realized makespan and $M^{*}$ be the oracle/reference makespan for the same instance.  The score is
\begin{equation}
  C_{\mathrm{D1}}=\min\left(1,\frac{M^{*}}{M}\right).
\end{equation}
A high score means the team finishes close to the reference parallel schedule; the score decreases when work is duplicated, imbalanced, or serialized.

\textbf{D2: Sequential ordering} covers procedures with causal order, such as opening a drawer or cabinet before placement and closing it only after all required objects are inside.  The coordination target is precedence compliance.  Let $n$ be the number of evaluated precedence-relevant operations, $v$ be the number of ordering-related violations, and $p_{\mathrm{legal}}$ be a fixed legality gate: $p_{\mathrm{legal}}=1$ if the trajectory contains no unauthorized skill use and no precondition violation, and $p_{\mathrm{legal}}=0.5$ otherwise. The score is
\begin{equation}
  C_{\mathrm{D2}}=p_{\mathrm{legal}}\left(1-\frac{v}{n}\right).
\end{equation}
A high score indicates compliance with the required order, while precondition violations lower the score.

\textbf{D3: Mutual exclusion} covers activities that require a shared exclusive tool, such as multiple agents preparing food with one knife.  The coordination target is contention-free resource scheduling.  Let $a$ be the number of shared-resource access attempts, $c$ be the number of access collisions, $L$ be the number of executed action steps, and $M^{*}$ be the oracle/reference makespan.  The score is
\begin{equation}
  C_{\mathrm{D3}}=\left(1-\frac{c}{a}\right)\min\left(1,\frac{M^{*}}{L}\right).
\end{equation}
The first term penalizes direct contention, and the second penalizes schedules that are substantially longer than the reference plan.

\textbf{D4: Handoff coordination} covers transport tasks where agents pass objects through an intermediate surface, such as a table or container.  The coordination target is paced producer-consumer behavior: upstream agents should not overflow the buffer, and downstream agents should not wait on an empty one.  Let $h$ be the number of successful buffer handoffs, $o$ be the number of overflow events, and $w$ be the number of empty-buffer waiting events.  The score is
\begin{equation}
  C_{\mathrm{D4}}=1-\min\left(1,\frac{o+w}{h}\right).
\end{equation}
A high score means producers and consumers maintain a compatible pace around the buffer.

\begin{figure*}[t]
  \centering
  \includegraphics[width=\linewidth]{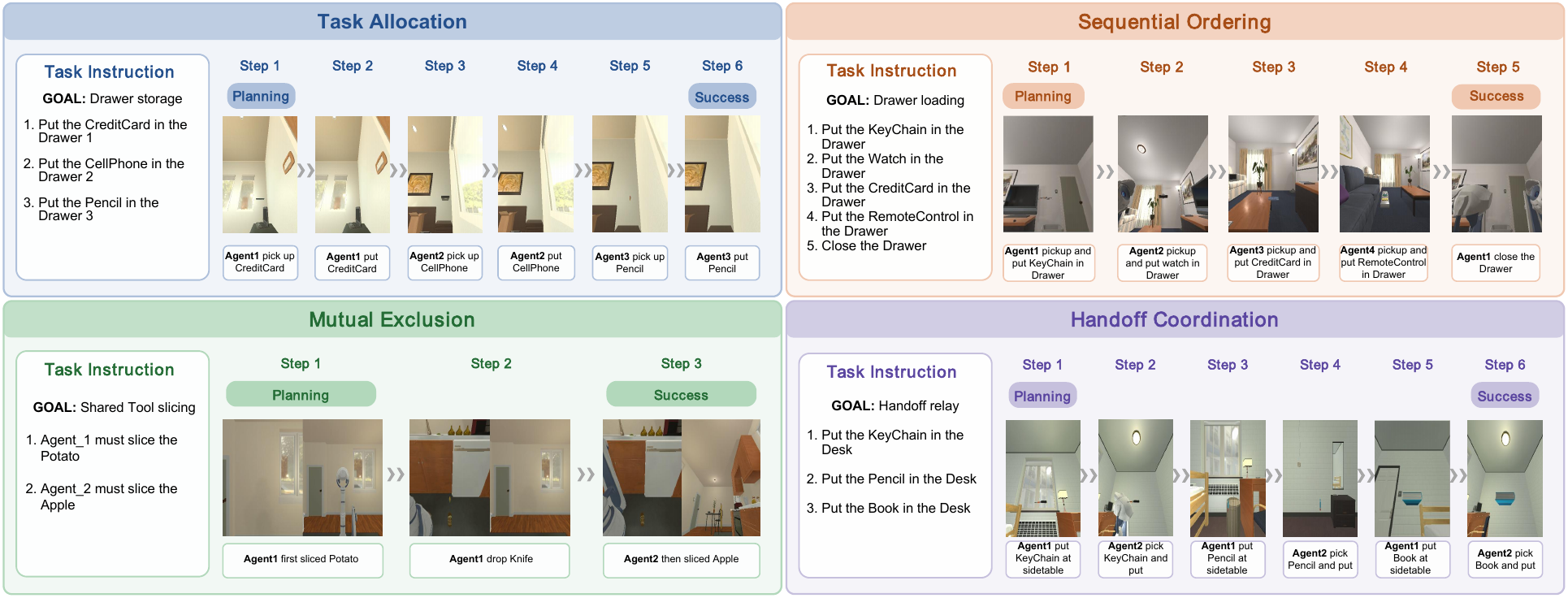}
  \caption{\textbf{Qualitative examples} across four coordination dimensions.}
  \label{fig:qualitative}
\end{figure*}

\subsection{Benchmark Instantiation}
\label{sec:task-instantiation}

We instantiate the four coordination constructs through household activities that naturally elicit their corresponding interaction patterns: object rearrangement for task allocation and sequential ordering, food preparation for shared-tool exclusion, and object transport for relay handoff. CoCoBench contains 897 validated instances spanning ten task families, four room categories, and three team sizes ($N\in\{2,3,4\}$). Instances are sampled from a task-family--construct matrix, retaining only combinations that provide executable and measurable coordination signals. Details of the benchmark design are provided in the Appendix.

All tasks share a fixed high-level skill interface covering navigation, waiting, object pickup and placement, container opening and closing, slicing, and other household interactions. At each step, agents are exposed only to parameterized skills relevant to the current task and state. The same executable interface also grounds construct-specific signals: container-state transitions reveal D2 ordering violations, exclusive-tool checks expose D3 contention, and role- and buffer-conditioned actions identify D4 handoff failures. 

We generate instances from symbolic templates grounded in simulator metadata. Each generated instance must pass three quality-control gates. First, a \textbf{placement gate} verifies object availability, spatial feasibility, and required affordances. Second, an \textbf{oracle gate} confirms task solvability using a privileged reference policy. Third, a \textbf{metric gate} retains only instances whose oracle trajectory achieves a near-ceiling score on the target construct. Together, these gates remove infeasible, ambiguous, and weakly diagnostic instances, prioritizing construct validity over raw data scale.

\subsection{Agent Evaluation Design}
\label{sec:agent-eval-pipeline}

CoCoBench evaluates high-level coordination policies by executing their selected skills and scoring the resulting trajectories. At step $t$, agent $a_i$ receives an observation $o_t^i=O_i(s_t)$ and an executable action menu
\begin{equation}
  \mathcal{U}_t^i
  =
  \Gamma(\Phi,r_i,s_t)
  \subseteq
  \mathcal{U},
\end{equation}
where $r_i$ is the assigned role and $\Gamma$ generates task-, role-, and state-conditioned actions. D1--D3 use homogeneous task-specific menus, whereas D4 assigns producer actions to source-to-buffer agents and consumer actions to buffer-to-target agents. All protocols use the same tasks, skill interface, and simulator environment.

A fixed executor applies the selected skills. In concurrent decentralized evaluation, agents select actions from the same state snapshot, which are then executed in a deterministic order. This preserves reproducibility while exposing conflicts such as competing for the same object, container, or exclusive tool.
The recorder logs states, observations, selected skills, execution outcomes, and failure events to produce a trajectory $\tau$. Metrics are computed deterministically from $\tau$:
\begin{equation}
\begin{aligned}
  \mathrm{SR}(\tau)
  &=
  \mathbf{1}\!\left[
  \forall \phi\in\Phi,\ \phi(s_T)=1
  \right], \\
  C_d(\tau)
  &=
  f_d(\tau),
  \qquad
  d\in\{\mathrm{D1},\mathrm{D2},\mathrm{D3},\mathrm{D4}\}.
\end{aligned}
\end{equation}
Here, $f_d$ denotes the construct-specific scoring function. By fixing the low-level executor, CoCoBench isolates differences in high-level team coordination.

\begin{table*}[t!]
\centering
\caption{\textbf{Main results on CoCoBench.} Results are reported on the full 897-instance set under the centralized image-observation protocol. \textbf{SR} = task success rate (\%); \textbf{CS} = construct score (0--1). Best per column in \textbf{bold}.}
\label{tab:exp1-main}
\resizebox{\textwidth}{!}{%
\begin{tabular}{@{}l*{10}{c}@{}}
\toprule
& \multicolumn{2}{c}{\textbf{Overall}} & \multicolumn{2}{>{\columncolor[HTML]{AEC0DB}}c}{\textbf{D1: Task allocation}} & \multicolumn{2}{>{\columncolor[HTML]{F0C8B1}}c}{\textbf{D2: Sequential ordering}} & \multicolumn{2}{>{\columncolor[HTML]{B2D8BB}}c}{\textbf{D3: Mutual exclusion}} & \multicolumn{2}{>{\columncolor[HTML]{C6C0DD}}c}{\textbf{D4: Handoff coordination}} \\
\cmidrule(lr){2-3} \cmidrule(lr){4-5} \cmidrule(lr){6-7} \cmidrule(lr){8-9} \cmidrule(lr){10-11}
Model & SR & CS & SR & CS & SR & CS & SR & CS & SR & CS \\
\midrule
\multicolumn{1}{@{}>{\columncolor[HTML]{F3F3F3}}l}{\textit{Proprietary (API)}} & \multicolumn{10}{>{\columncolor[HTML]{F3F3F3}}c@{}}{} \\
GPT-5.6-sol & \cellcolor[HTML]{D9D9D9}\textbf{84.8} & \cellcolor[HTML]{D9D9D9}\textbf{0.90} & 57.3 & 0.63 & 90.2 & 0.96 & \cellcolor[HTML]{DDEEE1}\textbf{96.5} & \cellcolor[HTML]{DDEEE1}\textbf{1.00} & \cellcolor[HTML]{E6E3F0}\textbf{94.2} & \cellcolor[HTML]{E6E3F0}\textbf{0.99} \\
GPT-5.4-mini & 33.6 & 0.51 & 81.7 & 0.66 & 30.7 & 0.60 & 10.9 & 0.59 & 12.9 & 0.18 \\
Claude Opus 4.8 & 78.5 & 0.87 & 38.1 & 0.50 & \cellcolor[HTML]{F8E6DC}\textbf{94.7} & \cellcolor[HTML]{F8E6DC}\textbf{1.00} & 93.9 & 0.99 & 85.7 & 0.99 \\
Claude Haiku 4.5 & 63.1 & 0.71 & 89.4 & 0.48 & 47.1 & 0.71 & 75.7 & 0.93 & 40.6 & 0.70 \\
Qwen3.6-Plus & 69.1 & 0.70 & 78.4 & 0.57 & 61.8 & 0.79 & 88.3 & 0.96 & 47.8 & 0.45 \\
Kimi-K2.6 & 57.6 & 0.71 & 71.6 & 0.64 & 3.6 & 0.44 & 93.5 & 0.99 & 61.6 & 0.77 \\
\midrule
\multicolumn{1}{@{}>{\columncolor[HTML]{F3F3F3}}l}{\textit{Open-weight}} & \multicolumn{10}{>{\columncolor[HTML]{F3F3F3}}c@{}}{} \\
Qwen3.5-9B & 36.7 & 0.63 & 91.3 & 0.61 & 37.8 & 0.65 & 17.0 & 0.77 & 2.7 & 0.48 \\
Qwen3-VL-8B & 24.6 & 0.57 & \cellcolor[HTML]{DBE3EF}\textbf{95.0} & \cellcolor[HTML]{DBE3EF}\textbf{0.91} & 4.9 & 0.45 & 1.3 & 0.73 & 0.0 & 0.18 \\
Cosmos3-Nano-16B & 29.5 & 0.75 & 79.8 & 0.89 & 37.8 & 0.64 & 1.7 & 0.86 & 0.9 & 0.44 \\
RoboBrain2.5-8B & 22.1 & 0.43 & 78.0 & 0.78 & 2.7 & 0.42 & 3.5 & 0.35 & 6.2 & 0.18 \\
InternVL3.5-8B & 13.4 & 0.42 & 53.7 & 0.45 & 1.3 & 0.41 & 0.0 & 0.56 & 0.0 & 0.26 \\
\bottomrule
\end{tabular}}
\end{table*}

\section{Experiments}
\label{sec:experiments}


\subsection{Experimental Setup}
\label{sec:experimental-setup}

\paragraph{Test Sets.} We evaluate models on two splits of CoCoBench. The full set contains 897 instances across four coordination constructs, four room categories, and three team sizes. For controlled ablations, we use a stratified 240-instance subset with 60 instances per construct and 20 instances for each construct $\times$ agent-count pair. This subset keeps the construct and team-size axes balanced while reducing the evaluation cost of architecture and observation ablations.

\paragraph{Metrics.} We report task success rate (SR) and construct score (CS). SR measures whether all goal predicates are satisfied at the end of the episode. CS measures whether the target coordination construct is handled effectively, using the construct-specific scores. We additionally track legality and terminal failure events, and use them for the diagnostic analyses. SR and CS are the primary summary metrics.

\paragraph{Models and Protocols.} The main comparison evaluates eleven MLLMs under the the full 897-instance set. Each episode uses the task-specific high-level skill step budget specified by the benchmark configuration, calibrated from oracle-validated trajectories; budget exhaustion is counted as failure and analyzed separately. The model suite includes proprietary API systems (GPT-5.6-sol, GPT-5.4-mini, Claude Opus 4.8, Claude Haiku 4.5, Qwen3.6-Plus, and Kimi-K2.6) and open-weight models (Qwen3.5-9B, Qwen3-VL-8B, Cosmos3-Nano-16B, RoboBrain2.5-8B, and InternVL3.5-8B). Among them, Cosmos3-Nano-16B~\cite{cosmos3} and RoboBrain2.5-8B~\cite{robobrain25} are open-weight models trained or adapted for robotic or embodied decision-making tasks. Full experimental details, parameter settings, and case studies are provided in the Appendix.

\subsection{Metric Validation}
\label{sec:metric-validation}

\begin{figure}[t]
  \centering
  \includegraphics[width=0.85\linewidth]{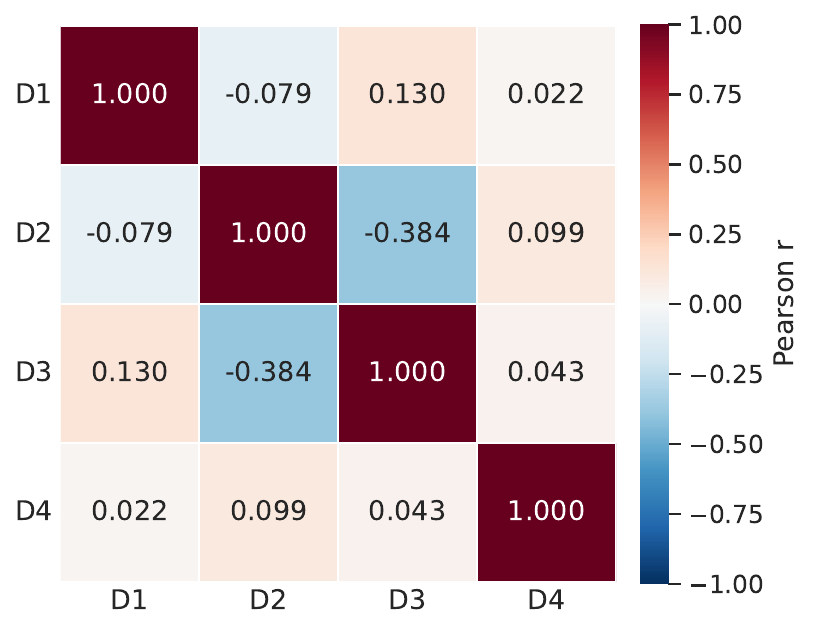}
   \caption{\textbf{Construct correlations.} Pairwise Pearson correlations between coordination scores on GPT-5.6 trajectories.}
   \label{fig:construct-correlation}
\end{figure}

We validate that $C_d$ measures four distinct coordination abilities rather than a single latent factor or a re-encoding of binary task success. Since each instance is scored only with respect to its designated target construct, we cross-apply all four scoring functions to the 897 GPT-5.6-sol trajectories from the centralized image-observation protocol, yielding a four-dimensional score vector for each trajectory. Figure~\ref{fig:construct-correlation} presents the resulting Pearson correlation matrix, with each pair computed over trajectories that provide sufficient evidence for both construct scores. All off-diagonal entries lie within $[-0.10, 0.16]$. These results indicate that the four coordination constructs are largely non-redundant.

CS is a trajectory-derived score that captures information unavailable from task success alone. Meanwhile, trajectory-level construct scores reflect the corresponding coordination abilities only within trajectories that provide sufficient evidence for evaluating those abilities. They may therefore differ from the final task success score. Nevertheless, this distinction makes CS more suitable for attributing failures to specific deficiencies in cooperative coordination.

\subsection{Benchmark Results}
\label{sec:benchmark-results}


\paragraph{Coordination ability varies across constructs.}
Figure~\ref{fig:qualitative} visualizes representative cases for the four coordination constructs, and Table~\ref{tab:exp1-main} reports the main comparison on the full 897-instance set. GPT-5.6-sol achieves the highest overall SR (84.8\%) and CS (0.90). The construct-level results reveal substantial variation across coordination types: Qwen3-VL-8B achieves the best D1 SR and CS, Claude Opus 4.8 leads both metrics on D2, and GPT-5.6-sol leads both metrics on D3 and D4. These results also highlight the complementary roles of the two metrics: SR measures task completion, whereas CS evaluates the construct-relevant coordination behavior observed along the trajectory. Consequently, a model may obtain a relatively high CS despite a low SR when its observed coordination decisions are effective but the episode remains incomplete. Overall, performance on one construct does not necessarily transfer to others, motivating construct-level evaluation alongside aggregate task success.

\begin{figure}[t!]
  \centering
  \includegraphics[width=\linewidth]{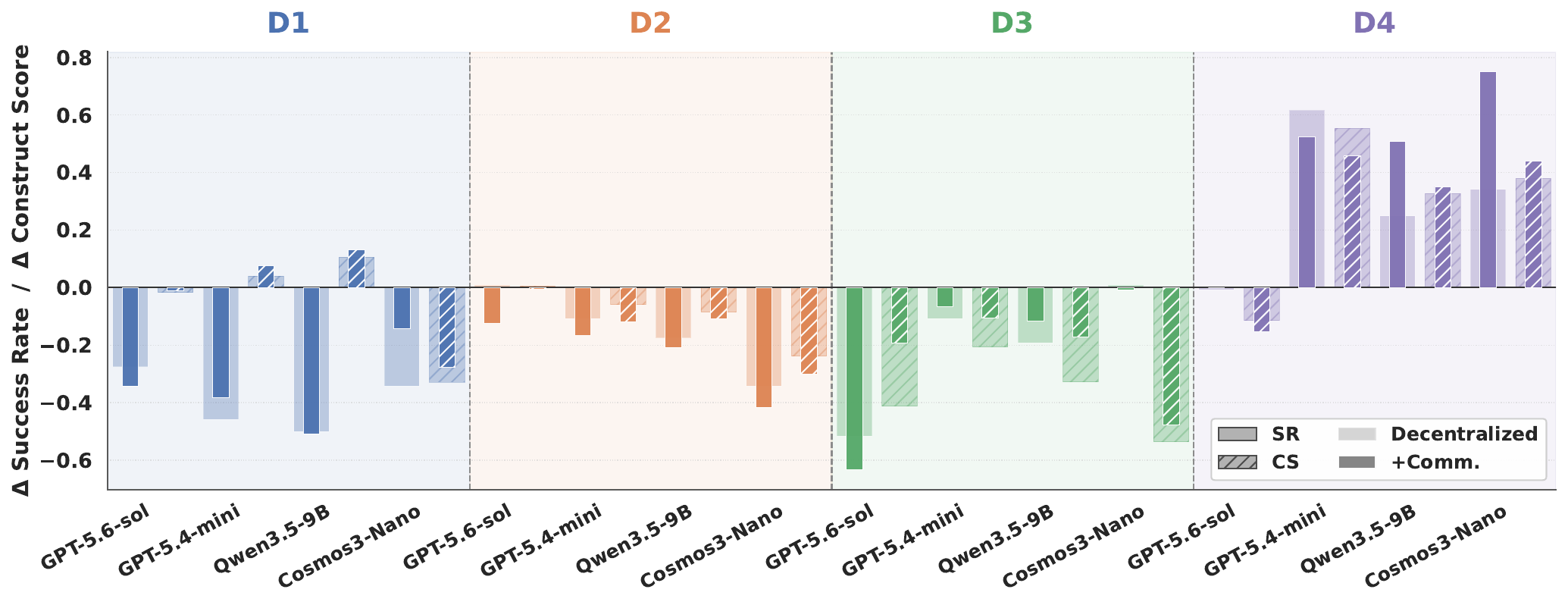}
  \caption{\textbf{Architecture effects on the 240-instance subset.} Signed change in SR
  and CS of the two decentralized policies relative to the centralized baseline,
  per model and construct, pooled over image and blind observations.}
  \label{exp2:architecture}
\end{figure}

\begin{figure}[t!]
  \centering
  \includegraphics[width=\linewidth]{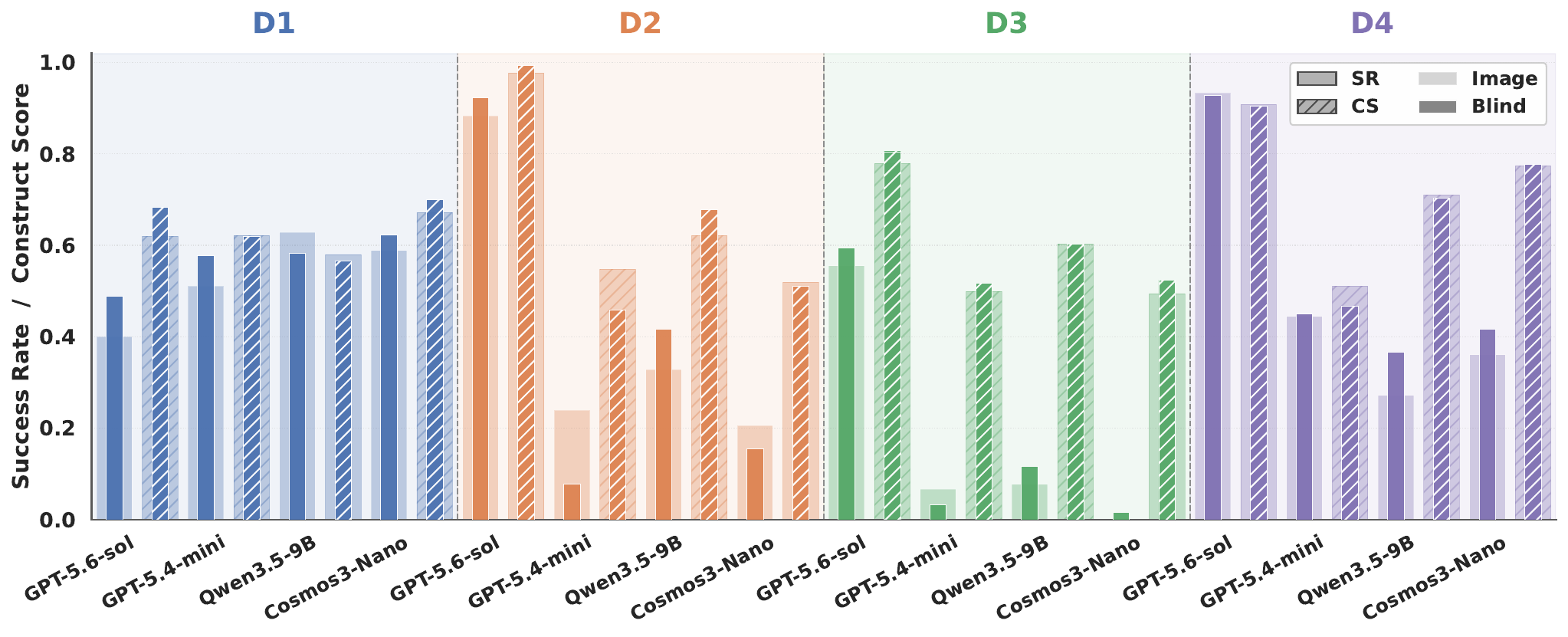}
  \caption{\textbf{Observation ablation on the 240-instance subset.} SR and CS under the image
  versus blind settings, per model and construct, pooled over the three policy
  architectures.}
  \label{exp3:vision}
\end{figure}

\paragraph{Centralized state aggregation improves coordination.} We examine whether coordination improves when the policy can aggregate team-level state before acting. Figure~\ref{exp2:architecture} compares three policy protocols on the balanced 240-instance subset. The centralized policy (C0) serves as the baseline, and the two decentralized variants are reported as signed changes relative to it. To isolate the effect of policy architecture, results are pooled over image and blind observations. C0 aggregates all agent observations and histories before selecting team actions, whereas decentralized planning gives each agent only its local observation and private history; the communication-enabled variant additionally provides a shared broadcast channel. Averaged across the four evaluated models and both observation settings, centralized planning achieves the best performance with 48.0\% SR and 0.69 CS, followed by decentralized planning with communication (39.6\% SR, 0.65 CS) and without communication (36.7\% SR, 0.63 CS). These results show that communication generally benefits decentralized coordination, but only partially compensates for the loss of centralized state aggregation.

\paragraph{Symbolic coordination is the primary bottleneck.} We then examine whether visual perception is the main limiting factor under CoCoBench's high-level skill interface. Figure~\ref{exp3:vision} compares the default image setting against a blind setting on the same balanced 240-instance subset and four models used for the architecture study. The image setting provides egocentric visual observations, while the blind setting removes images but preserves the textual task specification, action menu, and execution history. Removing images does not reduce aggregate performance: averaged across the four models and three policy architectures, blind evaluation obtains 42.3\% SR and 0.660 CS, compared with 40.6\% SR and 0.655 CS for image evaluation. Under CoCoBench's high-level skill interface, task-relevant objects and executable actions are already explicitly grounded, so the main bottleneck exposed by this benchmark is symbolic multi-agent coordination rather than raw visual recognition.

\begin{figure}[t]
  \centering
  \begin{subfigure}[t]{0.49\linewidth}
    \centering
    \includegraphics[width=\linewidth]{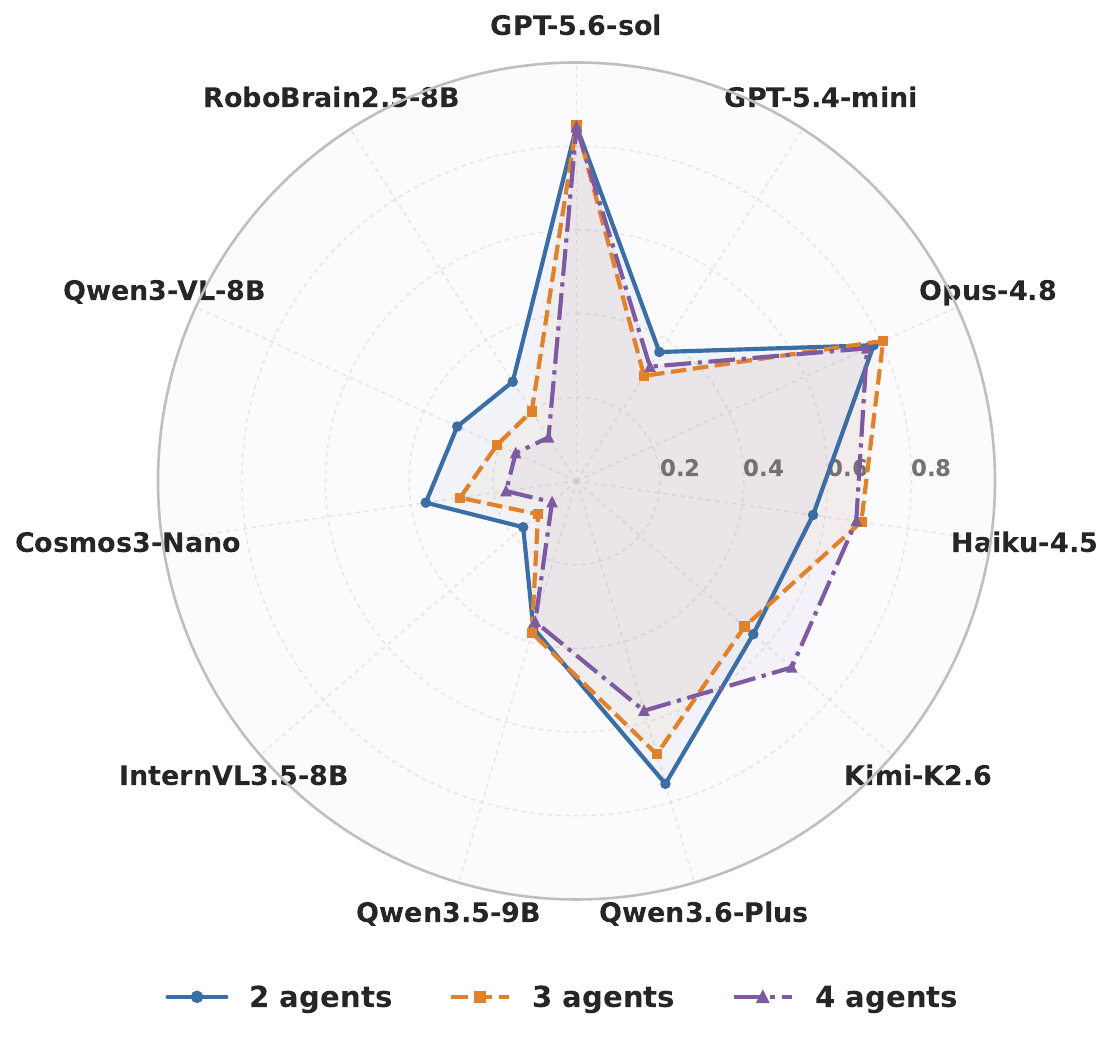}
    \caption{Success rate.}
    \label{exp5-1:agent-success}
  \end{subfigure}
  \hfill
  \begin{subfigure}[t]{0.49\linewidth}
    \centering
    \includegraphics[width=\linewidth]{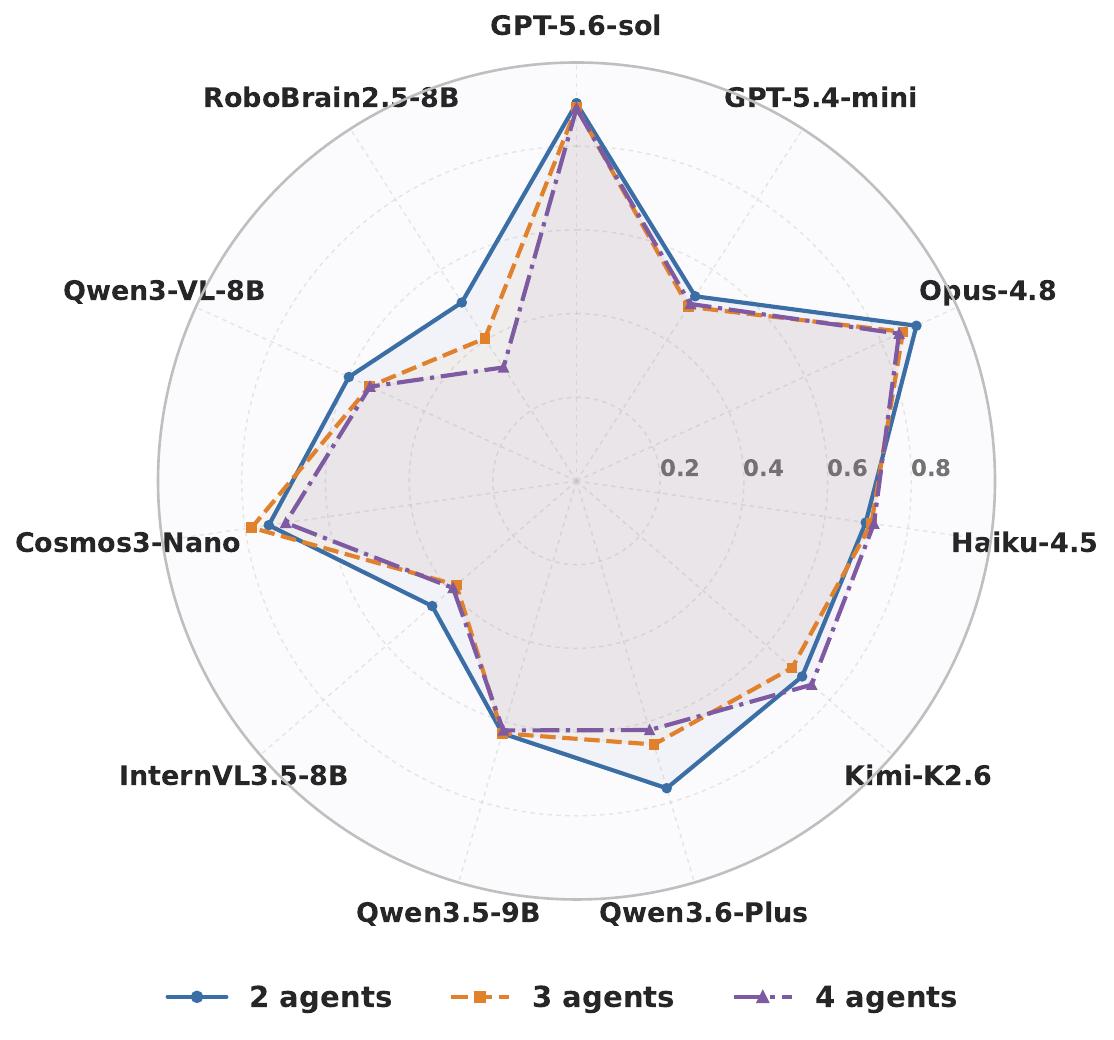}
    \caption{Construct score.}
    \label{exp5-2:agent-construct}
  \end{subfigure}
  \caption{\textbf{Scaling with team size.} Per-model SR (a) and CS (b) across two-,
  three-, and four-agent instances.}
  \label{exp5:agent-scaling}
\end{figure}

\paragraph{Coordination becomes harder as team size grows.} We ask whether adding agents merely increases the number of actions or instead amplifies the coordination burden. Figure~\ref{exp5-1:agent-success} and Figure~\ref{exp5-2:agent-construct} show performance across two-, three-, and four-agent instances. Averaged over all evaluated models, SR decreases from 48.8\% at two agents to 45.8\% at three agents and 43.1\% at four agents; CS similarly decreases from 0.68 to 0.64 and 0.63. The average decline is modest but systematic, and it masks a stronger model-dependent pattern. Strong models such as GPT-5.6-sol and Claude Opus 4.8 remain comparatively stable across team sizes, whereas weaker open-weight models degrade more sharply. For example, InternVL3.5-8B drops from 16.9\% SR at two agents to 7.7\% at four agents. These results suggest that adding agents increases coordination load even when the primitive skill interface remains fixed.

\begin{figure}[t]
  \centering
  \begin{subfigure}[t]{0.49\linewidth}
    \centering
    \includegraphics[width=\linewidth]{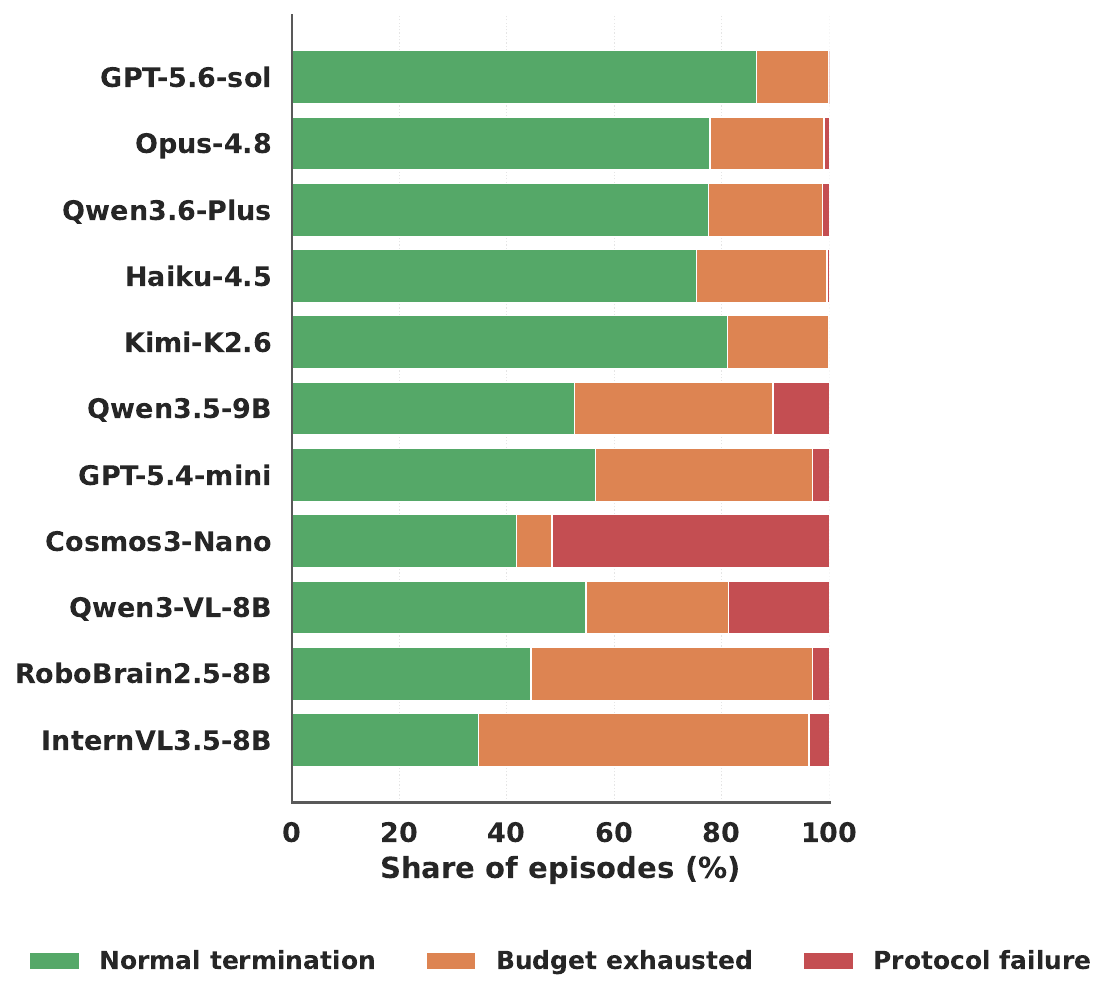}
    \caption{Failure modes.}
    \label{exp6-1}
  \end{subfigure}
  \hfill
  \begin{subfigure}[t]{0.49\linewidth}
    \centering
    \includegraphics[width=\linewidth]{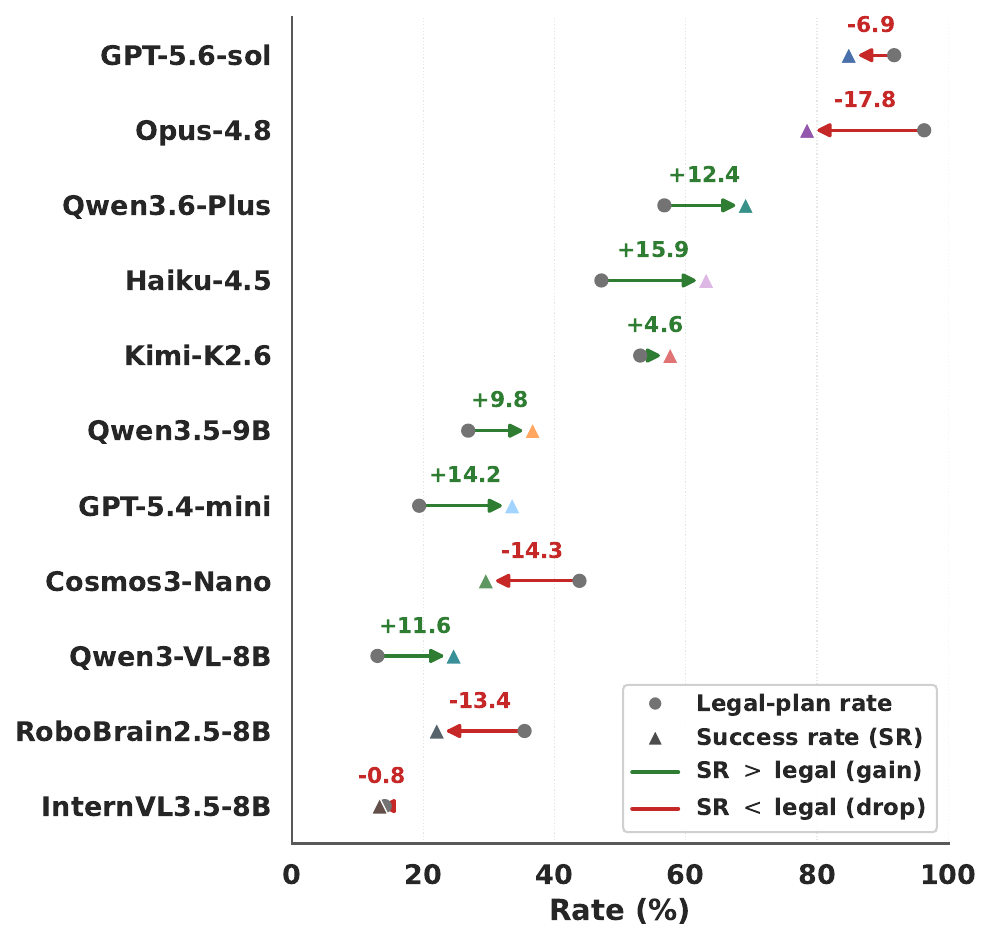}
    \caption{Legality gap.}
    \label{exp6-2}
  \end{subfigure}
  \caption{\textbf{Failure modes and legality.} Per-model terminal-state taxonomy (a)
  and success rate versus legal-plan rate (b).}
  \label{exp6:failure-legality}
\end{figure}

\subsection{Failure and Legality Analysis}
\label{sec:failure-analysis}

\paragraph{Failure patterns reveal distinct model bottlenecks.} Figure~\ref{exp6:failure-legality}(a) decomposes how episodes terminate. For proprietary models, protocol failures are essentially absent (mean $1.0\%$), so most unsolved episodes reflect normal termination or budget exhaustion. For open-weight models, by contrast, protocol failures average $17.5\%$, while budget exhaustion reaches $61\%$ for InternVL3.5-8B and $52\%$ for RoboBrain2.5-8B. Combined, budget-plus-protocol terminations account for $54.3\%$ of open-weight episodes on average. Therefore, part of the gap comes from unstable adherence to the high-level action protocol and difficulty completing planning and execution within the given step budget.

\paragraph{Task success and legal coordination diverge.} Figure~\ref{exp6:failure-legality}(b) shows why success alone is insufficient. Success and legality diverge, and the sign of the gap is informative. Claude Opus 4.8 produces legal plans on $96.3\%$ of episodes but succeeds on $78.5\%$, while GPT-5.6-sol is legal on $91.8\%$ but succeeds on $84.8\%$: both often honor the coordination constraints even when they ultimately miss the goal. Conversely, several models achieve higher success than legality---Claude Haiku 4.5 ($63.1\%$ SR vs.\ $47.2\%$ legal), Qwen3.6-Plus ($69.1\%$ vs.\ $56.7\%$), and Qwen3-VL-8B ($24.6\%$ vs.\ $13.0\%$)---so their binary success overstates how well they actually coordinate. Because a goal predicate can sometimes be reached by an illegal shortcut, legality and CS are needed to separate ``coordinates correctly'' from ``reaches the goal by any means.''

\section{Conclusion}
\label{sec:conclusion}
We introduced \textbf{CoCoBench}, a construct-level benchmark for evaluating multi-agent embodied coordination in executable household environments. By organizing tasks around four dimensions, CoCoBench separates goal completion from coordination quality and enables judge-free, construct-level diagnosis.
Our evaluation of 11 leading MLLMs shows that coordination ability is highly structured: strong overall performance does not imply balanced competence across coordination constructs, centralized state aggregation remains important, and increasing team size amplifies coordination difficulty. We hope CoCoBench will support more fine-grained evaluation and inspire future models that reason explicitly about collaborative execution.

\bibliography{aaai2027}

\appendix
\twocolumn[
  \begin{center}
    {\LARGE\bfseries Appendix}
  \end{center}
]

\section{Benchmark Details}

This section presents additional benchmark-side details of CoCoBench. We describe how the four coordination constructs are realized as household tasks, how symbolic templates are grounded and validated in AI2-THOR, the composition of the resulting benchmark, the agent inputs, and representative successful trajectories.

\subsection{Task Definition}

Table~\ref{tab:task-definitions} lists the ten task types in CoCoBench. Each task type is tied to one coordination dimension and instantiated with scene-specific object names, receptacles, and agent counts. The four dimensions cover task allocation, sequential ordering, mutual exclusion, and handoff coordination.

\paragraph{D1: Task Allocation.}
This dimension evaluates whether a team can decompose a conjunctive goal into independent units of work and assign them without unnecessary duplication. In \emph{Store Groceries}, agents may concurrently transport different food items while another agent handles the light switch. \emph{Sort \& Store} and \emph{Tidy Drawers} expose several object--destination pairs that can be completed in parallel. Although any agent could execute multiple pairs, effective collaboration requires a balanced division that reduces idle time and approaches the oracle makespan. Thus, final-state success alone is insufficient: a successful but serialized trajectory does not demonstrate intended allocation behavior.

\paragraph{D2: Sequential Ordering.}
This dimension introduces causal dependencies between otherwise distributable subtasks. In \emph{Stock Cabinet}, \emph{Load in Order}, and \emph{Fill Drawers in Order}, agents must first make the target receptacle accessible, complete all required placements, and only then close it. Parallel transport is useful, but premature closing invalidates or blocks a teammate's pending placement. These tasks therefore test whether agents jointly maintain a shared notion of progress and delay terminal actions until their preconditions have been satisfied.

\paragraph{D3: Mutual Exclusion.}
This dimension evaluates contention-free scheduling around a shared exclusive resource. In \emph{Prepare Food}, multiple agents are assigned different sliceable objects, but only one Knife is available. An agent must acquire the Knife, complete its assigned slicing operation, and release or transfer access so that the next agent can proceed. Simultaneous acquisition attempts, repeated searches for an already-held tool, and retaining the Knife after use create avoidable contention. The task separates ordinary object manipulation from the ability to coordinate ownership of a scarce resource.

\paragraph{D4: Handoff Coordination.}
This dimension covers transport tasks whose completion requires objects to cross an intermediate buffer or spatial boundary. \emph{Relay Items} and \emph{Indirect Relay} use a shared surface, plate, or container as a transfer point, whereas \emph{Cross-Zone Handoff} assigns agents to separated source and destination regions. Upstream agents act as producers that deliver objects to the buffer, while downstream agents act as consumers that carry them to the final receptacle. Success requires compatible pacing: producers should avoid overflowing or contending for the transfer point, and consumers should avoid waiting at an empty buffer. The trajectories expose whether agents can sustain role-aware cooperation over handoff cycles.

\begin{table*}[!t]
\centering
\caption{Definitions and instruction examples of ten coordination-oriented task types in CoCoBench.}
\label{tab:task-definitions}
\begingroup
\footnotesize
\setlength{\tabcolsep}{3pt}
\begin{tabular}{@{}m{0.135\textwidth}m{0.125\textwidth}m{0.405\textwidth}m{0.265\textwidth}@{}}
\toprule
{\centering\textbf{Coordination Type}\par} &
{\centering\textbf{Task Type}\par} &
{\centering\textbf{Instruction Example}\par} &
{\centering\textbf{Description}\par} \\
\midrule

&
{\centering Store Groceries\par} &
{\raggedright\itshape Put the Potato and Lettuce in the Fridge, and turn off the LightSwitch.\par
Put the Potato in the Fridge, put the Lettuce in the Fridge, and make sure the LightSwitch is off.\par} &
{\raggedright Agents divide independent household subtasks, such as storing food and switching off a light, and finish all objectives without repeated actions.\par} \\
\cmidrule(lr){2-4}

{\centering\textbf{D1}\\\textbf{Task Allocation}\par} &
{\centering Sort \& Store\par} &
{\raggedright\itshape Put the CreditCard on the CoffeeTable, the CellPhone on the Desk, and the Pencil on the ArmChair.\par} &
{\raggedright Agents sort multiple personal objects to designated receptacles and split object-target pairs across the team.\par} \\
\cmidrule(lr){2-4}

&
{\centering Tidy Drawers\par} &
{\raggedright\itshape Put the CreditCard in Drawer\_1, the CellPhone in Drawer\_2, and the Pencil in Drawer\_3.\par} &
{\raggedright Agents place personal items into assigned drawers, opening drawers before placement when needed.\par} \\
\midrule

&
{\centering Stock Cabinet\par} &
{\raggedright\itshape Put the Apple and Lettuce in the Cabinet, then close the Cabinet after both items are inside.\par} &
{\raggedright Agents load kitchen items into a cabinet and close it only after all required items have been placed.\par} \\
\cmidrule(lr){2-4}

{\centering\textbf{D2}\\\textbf{Sequential Ordering}\par} &
{\centering Load in Order\par} &
{\raggedright\itshape Put the KeyChain, Watch, CreditCard, and RemoteControl in the Box, then close the Box.\par} &
{\raggedright Agents load several objects into a container and perform the final close action after the loading sequence is complete.\par} \\
\cmidrule(lr){2-4}

&
{\centering Fill Drawers in Order\par} &
{\raggedright\itshape Put the KeyChain, Watch, CreditCard, and RemoteControl in the Drawer, then close the Drawer.\par} &
{\raggedright Agents place several personal objects into one drawer and close it only after all placements are complete.\par} \\
\midrule

{\centering\textbf{D3}\\\textbf{Mutual Exclusion}\par} &
{\centering Prepare Food\par} &
{\raggedright\itshape agent\_1 must slice the Potato, agent\_2 must slice the Apple, and agent\_3 must slice the Tomato.\par} &
{\raggedright Agents prepare food with a shared Knife and avoid blocking teammates from using the shared tool.\par} \\
\midrule

&
{\centering Relay Items\par} &
{\raggedright\itshape Move the CellPhone, Pen, and Book to the Desk by relaying them through the intermediate SideTable.\par} &
{\raggedright Agents use an intermediate receptacle as a handoff point and coordinate producer and consumer roles.\par} \\
\cmidrule(lr){2-4}

{\centering\textbf{D4}\\\textbf{Handoff Coordination}\par} &
{\centering Indirect Relay\par} &
{\raggedright\itshape Put the Potato, Tomato, and Apple in the Fridge by first using the Plate as a transfer point.\par} &
{\raggedright Agents pass food objects through an intermediate carrier or receptacle before the final placement.\par} \\
\cmidrule(lr){2-4}

&
{\centering Cross-Zone Handoff\par} &
{\raggedright\itshape Relay the KeyChain, Pencil, and Book across room zones so they end up on the Desk.\par} &
{\raggedright Agents cooperate across separated zones by delivering objects to a handoff location and then to the final target.\par} \\
\bottomrule

\end{tabular}
\endgroup
\end{table*}

\subsection{Benchmark Construction Details}

CoCoBench instances are produced by grounding symbolic coordination templates in executable AI2-THOR scenes. Each template specifies a room category, object classes, goal predicates, team size, target coordination construct, and construct-specific constraints. The generator samples a compatible floorplan, concrete object instances, legal initial placements, agent roles where applicable, and a random seed. Natural-language instructions are then rendered from the grounded goal predicates so that all named objects and receptacles correspond to entities available in the scene.

The executable interface is held fixed across instances. At each step, scene metadata and task constraints are used to construct parameterized high-level skills for navigation, waiting, pickup, placement, opening and closing, and related interactions. D1--D3 expose homogeneous task-bound menus, whereas D4 uses role-conditioned menus that distinguish source-to-buffer producer actions from buffer-to-target consumer actions. All outcomes are determined by simulator execution and recorded together with observations, selected skills, state transitions, and failures; benchmark scores can therefore be reproduced from the trajectory.

\begin{figure}[t]
  \centering
  \includegraphics[width=\linewidth]{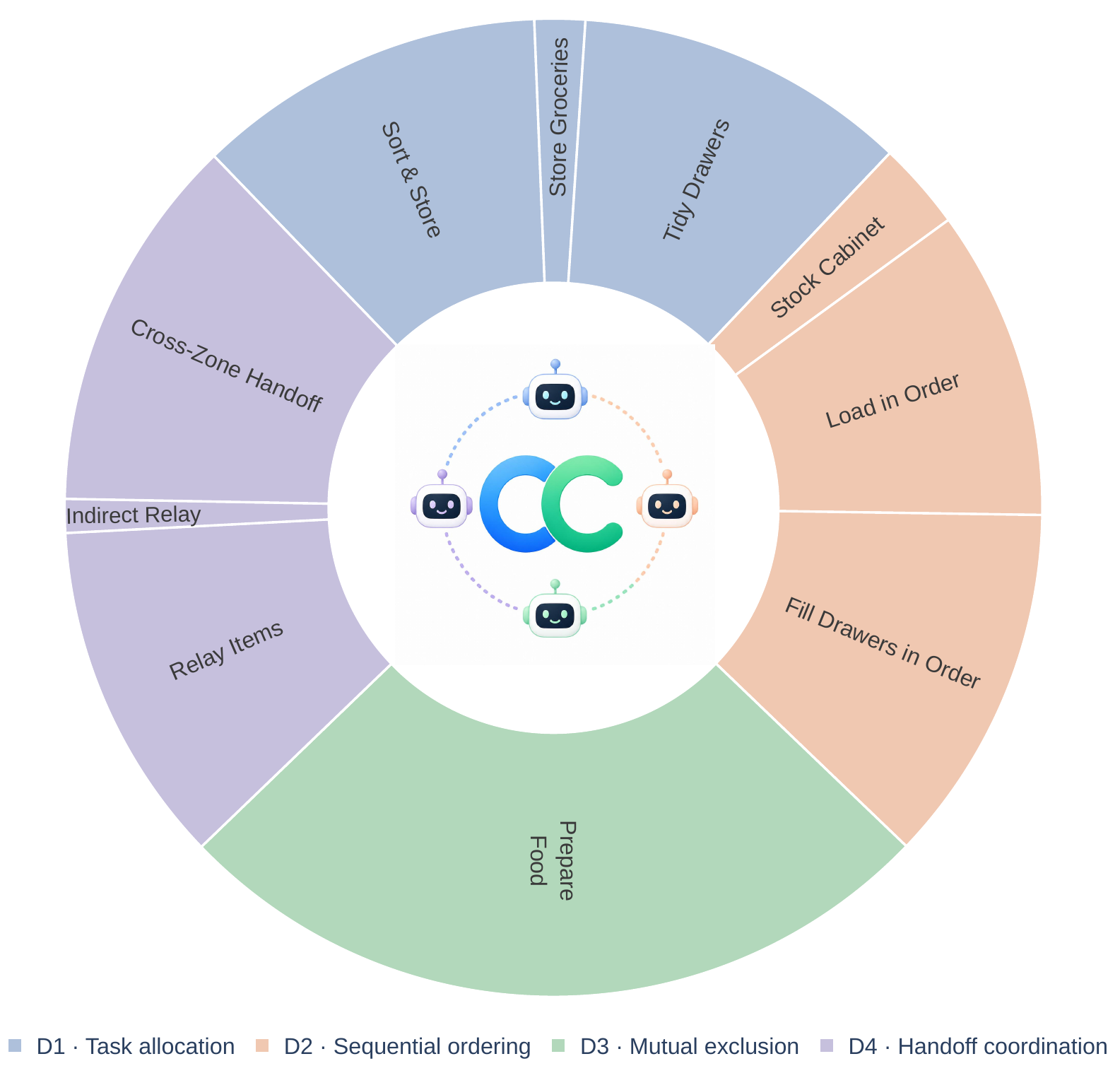}
  \caption{CoCoBench composition across coordination constructs and task families.}
  \label{fig:task-sunburst}
\end{figure}

\subsection{Benchmark Statistics}

The full benchmark contains \textbf{897 oracle-validated instances} spanning four coordination dimensions, six household task families, four room categories, and three team sizes as shown in Figure~\ref{fig:task-sunburst}. The construct distribution is nearly balanced: D1 task allocation contains 218 instances, D2 sequential ordering 225, D3 mutual exclusion 230, and D4 handoff coordination 224. By team size, the benchmark contains 415 two-agent, 288 three-agent, and 194 four-agent instances. This composition evaluates the same coordination demand under increasing team sizes while avoiding domination by any single construct. For controlled architecture and observation ablations, the main paper additionally uses a stratified 240-instance subset with 60 instances per construct and 20 instances for every construct--team-size pair. 

\subsection{Oracle Reference Plans}

Oracle plans are constructed from manually written privileged plan templates, one for each task--dimension pair. We implement 12 oracle templates and instantiate them into 897 per-instance oracle plans. Each template specifies the grounded subgoals assigned to each agent and the required order of high-level skills, and is instantiated using the scene, team size, roles, and ground-truth object IDs in the task configuration. D1 templates balance independent subtasks across agents; D2 templates encode precedence constraints; D3 templates serialize ownership of shared resources; and D4 templates explicitly schedule producer--buffer--consumer transfers. For distributed validation, the central oracle runs in the concurrent runner: in each round, only the agent designated by the oracle acts, while other agents remain idle, preserving the central reference trace.

The oracle's optimality is limited to the high-level skill abstraction. For D1, we manually verify that the reference makespan $M^{\ast}$ is optimal under this abstraction and use $M^{\ast}$ to compute the D1 construct score. For D2--D4, we do not claim that the oracle trajectories are globally shortest in either high-level skill count or low-level navigation.

Recoverable single-skill failures are handled by retrying or switching execution branches within the oracle script. If the final evaluation fails, the instance is repaired, regenerated, or removed. Only instances with a successful final state and a legal plan are included in the benchmark. We use this procedure to verify the executability of all current tasks and establish their theoretical upper bound.

\section{Evaluation Setup}

This section lists the evaluated model set and the action space exposed to the planner. The simulator uses metadata to build action-id menus; the model selects an action id, and the executor resolves it to a concrete skill call.

\subsection{Model Versions}

Table~\ref{tab:model-names} lists the proprietary and open-weight MLLMs used in our evaluation. We report the short names used in figures together with the corresponding provider or model family.

\begin{table*}[!t]
\centering
\caption{Models used in our evaluation.}
\label{tab:model-names}
\small
\setlength{\tabcolsep}{5pt}
\begin{tabular}{@{}p{0.27\textwidth}p{0.18\textwidth}p{0.47\textwidth}@{}}
\toprule
\textbf{Model Name} & \textbf{Creator} & \textbf{Full Name / Identifier} \\
\midrule
GPT-5.6-sol & OpenAI & gpt-5.6-sol \\
GPT-5.4-mini & OpenAI & gpt-5.4-mini \\
Claude Opus 4.8 & Anthropic & claude-opus-4.8 \\
Claude Haiku 4.5 & Anthropic & claude-haiku-4.5 \\
Qwen3.6-Plus & Qwen & qwen3.6-plus \\
Kimi-K2.6 & Moonshot AI & kimi-k2.6 \\
\midrule
Qwen3.5-9B & Qwen & Qwen3.5-9B \\
Qwen3-VL-8B & Qwen & Qwen3-VL-8B-Instruct \\
Cosmos3-Nano-16B & NVIDIA & NVIDIA Cosmos 3 Nano \\
RoboBrain2.5-8B & BAAI / RoboBrain Team & RoboBrain2.5-8B-NV \\
InternVL3.5-8B & OpenGVLab & InternVL3.5-8B-Instruct \\
\bottomrule
\end{tabular}
\end{table*}

\subsection{Action Space}

Table~\ref{tab:skill-set} summarizes the skill-level action space. Repeated instances of the same object category are merged for readability; during execution, every menu entry is grounded to a concrete AI2-THOR object id.

\begin{table*}[!t]
\centering
\caption{Action types and target object categories exposed to the planner.}
\label{tab:skill-set}
\begingroup
\footnotesize
\setlength{\tabcolsep}{4pt}
\begin{tabular}{@{}m{0.200\textwidth}m{0.740\textwidth}@{}}
\toprule
{\centering\textbf{Action Type}\par} & {\centering\textbf{Target Object}\par} \\
\midrule
{\centering Find\par} & {\raggedright AlarmClock, Apple, ArmChair, BasketBall, Bathtub, Bed, Book, Box, Bread, ButterKnife, Cabinet, Candle, Cart, CellPhone, CD, Chair, Cloth, CoffeeMachine, CoffeeTable, CounterTop, CreditCard, Cup, Desk, DeskLamp, DishSponge, DiningTable, Dresser, Drawer, Egg, Faucet, FloorLamp, Fork, Fridge, GarbageCan, GlassBottle, HandTowel, Kettle, KeyChain, Knife, Ladle, Laptop, Lettuce, LightSwitch, Microwave, Mug, Newspaper, Ottoman, Pan, PepperShaker, Pencil, Pen, Pillow, Plate, Plunger, Potato, RemoteControl, Safe, SaltShaker, Shelf, SideTable, Sink, SoapBar, SoapBottle, Sofa, Spatula, SprayBottle, Statue, StoveBurner, TennisRacket, TissueBox, Tomato, Toilet, ToiletPaper, ToiletPaperHanger, Vase, Watch, WateringCan, WineBottle\par} \\
\midrule
{\centering Explore\par} & {\raggedright forward, back, left, right, turn\_left, turn\_right, look\_up, look\_down\par} \\
\midrule
{\centering Wait\par} & {\raggedright None\par} \\
\midrule
{\centering PickUp\par} & {\raggedright AlarmClock, Apple, BaseballBat, BasketBall, Book, Bowl, Box, Bread, ButterKnife, Candle, CD, CellPhone, Cloth, CreditCard, Cup, DishSponge, Egg, Fork, GlassBottle, HandTowel, Kettle, KeyChain, Knife, Ladle, Laptop, Lettuce, Mug, Newspaper, Pan, Pen, Pencil, PepperShaker, Plate, Plunger, Potato, RemoteControl, SaltShaker, SoapBar, SoapBottle, Spatula, SprayBottle, Spoon, Statue, TennisRacket, TissueBox, Tomato, Vase, Watch, WateringCan, WineBottle\par} \\
\midrule
{\centering Put\par} & {\raggedright ArmChair, Bed, Box, Cabinet, Chair, CoffeeTable, CounterTop, Desk, DiningTable, Drawer, Dresser, Fridge, Microwave, Plate, Safe, Shelf, SideTable, Sink, Sofa\par} \\
\midrule
{\centering Drop\par} & {\raggedright Object in hand\par} \\
\midrule
{\centering Open\par} & {\raggedright Box, Cabinet, Drawer, Fridge, Laptop, Microwave, Safe\par} \\
\midrule
{\centering Close\par} & {\raggedright Box, Cabinet, Drawer, Fridge, Laptop, Microwave, Safe\par} \\
\midrule
{\centering ToggleOn\par} & {\raggedright DeskLamp, Faucet, FloorLamp, LightSwitch, Microwave\par} \\
\midrule
{\centering ToggleOff\par} & {\raggedright DeskLamp, Faucet, FloorLamp, LightSwitch, Microwave\par} \\
\midrule
{\centering Slice\par} & {\raggedright Apple, Bread, Lettuce, Potato, Tomato\par} \\
\midrule
{\centering CleanObject\par} & {\raggedright Apple, Bowl, Cloth, Cup, DishSponge, Egg, Fork, Kettle, Knife, Lettuce, Mug, Pan, Plate, Potato, SoapBottle, Spatula, Tomato, Vase\par} \\
\midrule
{\centering FillObjectWith\\Liquid\par} & {\raggedright Bowl, Cup, Kettle, Mug, Pan, Plate, Pot, Vase\par} \\
\midrule
{\centering EmptyLiquidFrom\\Object\par} & {\raggedright Bowl, Cup, Kettle, Mug, Pan, Plate, Pot, Vase\par} \\
\midrule
{\centering PushObject /\\PullObject\par} & {\raggedright ArmChair, BasketBall, Box, Chair, CoffeeTable, DiningTable, Dresser, GarbageCan, Ottoman, Pillow, SideTable, Sofa, Statue\par} \\
\midrule
{\centering BreakObject\par} & {\raggedright Bottle, Bowl, CellPhone, Cup, DishSponge, Egg, GlassBottle, Laptop, Mirror, Mug, Plate, Statue, Television, Vase, Window\par} \\
\bottomrule
\end{tabular}
\endgroup
\end{table*}

\section{Supplementary Experiments}
\label{sec:supp-experiments}

This appendix reports two additional analyses on the full 897-instance, centralized, image-observation setting, using fields already recorded in every evaluation trajectory but not aggregated in the main text.

\subsection{Diagnostic Decomposition of Failures}
\label{sec:supp-failure-diagnostics}

Main text shown that failures can be budget-driven, protocol-driven, or coordination-driven, but reports these only in aggregate. Table~\ref{tab:supp-failure-diagnostics} breaks this down per model: for every failed episode we log whether it involved a dependency violation, an occupancy conflict, an affordance failure, or an illegal skill call (categories are not mutually exclusive), pooling D1--D4 by each dimension's failure count.

Dependency violations dominate failures for every model (11--95\%), while occupancy conflicts (1--7\%) and affordance failures (0--3\%) are rare, and illegal-skill use is never a cause of failure (0\% for all 11 models): the action menu restricts calls to task-relevant skills, so illegal attempts are blocked before they can propagate into an episode failure. Claude Opus 4.8 is the outlier at 11\%, the lowest dependency-violation share of any model, consistent with its 96.3\% legal-plan rate reported in main text: most of its failures are incomplete-but-legal rather than order-violating.

Task progress on failed episodes (Prog.\ fail) also separates models that fail early from models that fail near the goal: Cosmos3-Nano-16B collapses earliest (25.0\%), while Kimi-K2.6 reaches the highest partial progress before failing (65.8\%); GPT-5.6-sol reaches 42.0\% progress on its failed episodes. The invalid-action ratio cleanly separates the two model classes: every proprietary model stays under 8\%, while every open-weight model exceeds 14\%, giving per-model quantitative support to the protocol-adherence gap discussed qualitatively.

\begin{table}[t]
\centering
\caption{\textbf{Diagnostic decomposition of failed episodes} (full 897-set, centralized, image observation). \textbf{Prog.\ (fail)} = mean task progress on failed episodes (\%); \textbf{Steps (succ.)} = mean executed steps on successful episodes; \textbf{Inv.\ act.} = invalid-action ratio (\%); \textbf{Dep/Occ/Aff/Ill} = share of failed episodes exhibiting a dependency violation, occupancy conflict, affordance failure, or illegal-skill use (\%), pooled over D1--D4 and weighted by each dimension's failure count.}
\label{tab:supp-failure-diagnostics}
\resizebox{\linewidth}{!}{%
\begin{tabular}{l ccc cccc}
\toprule
Model & Prog.\ (fail) & Steps (succ.) & Inv.\ act. & Dep & Occ & Aff & Ill \\
\midrule
\multicolumn{8}{l}{\textit{Proprietary (API)}} \\
GPT-5.6-sol & 42.0 & 15.8 & 1.8 & 29 & 7 & 2 & 0 \\
GPT-5.4-mini & 45.0 & 16.7 & 0.0 & 95 & 6 & 3 & 0 \\
Claude Opus 4.8 & 39.8 & 16.7 & 1.7 & 11 & 7 & 0 & 0 \\
Claude Haiku 4.5 & 49.7 & 17.3 & 7.6 & 57 & 3 & 1 & 0 \\
Qwen3.6-Plus & 46.4 & 17.9 & 0.0 & 61 & 6 & 2 & 0 \\
Kimi-K2.6 & 65.8 & 18.0 & 6.3 & 74 & 7 & 1 & 0 \\
\midrule
\multicolumn{8}{l}{\textit{Open-weight}} \\
Qwen3.5-9B & 39.0 & 14.4 & 15.7 & 90 & 4 & 1 & 0 \\
Qwen3-VL-8B & 39.4 & 11.2 & 23.0 & 92 & 1 & 1 & 0 \\
Cosmos3-Nano-16B & 25.0 & 10.4 & 14.2 & 62 & 2 & 2 & 0 \\
RoboBrain2.5-8B & 43.6 & 11.2 & 19.4 & 75 & 7 & 2 & 0 \\
InternVL3.5-8B & 33.8 & 14.6 & 18.9 & 90 & 5 & 2 & 0 \\
\bottomrule
\end{tabular}}
\end{table}

\subsection{Per-Episode Compute Cost}
\label{sec:supp-efficiency}

Table~\ref{tab:supp-efficiency} reports the mean wall-clock time, executed action steps, and planner (MLLM) calls per episode for each model. Executed steps and planner calls track closely for every model (e.g.\ 18.4 vs.\ 19.2 for GPT-5.6-sol, 31.1 vs.\ 31.7 for GPT-5.4-mini), so most planner outputs execute directly rather than being discarded as invalid actions.

Cost and step-efficiency are largely decoupled across models. GPT-5.6-sol attains the highest SR (84.8\%) and the fewest average steps among proprietary models (18.4), at 123.8s per episode. Claude Opus 4.8 is substantially slower (296.9s) despite a comparable trajectory length, whereas Qwen3.6-Plus reaches a mid-tier SR (69.1\%) at the lowest cost among proprietary models (106.7s). RoboBrain2.5-8B and GPT-5.4-mini require substantially more steps per episode (32.1 and 31.1) while running at comparable or lower wall-clock cost (42.4s and 120.0s). Wall-clock time reflects our evaluation hardware and API latency rather than FLOPs, so it is informative for comparing models within the proprietary or within the open-weight group, but not for comparing across the two.

\begin{table}[t]
\centering
\caption{\textbf{Per-episode compute cost} (full 897-set, centralized, image observation). \textbf{Time} = mean wall-clock time per episode (s); \textbf{Steps} = mean executed action steps; \textbf{Planner calls} = mean planner (LLM) invocations per episode. Time reflects our evaluation hardware and API latency, not FLOPs, and is not directly comparable across proprietary and self-hosted open-weight models.}
\label{tab:supp-efficiency}
\begin{tabular}{l ccc}
\toprule
Model & Time (s) & Steps & Planner calls \\
\midrule
\multicolumn{4}{l}{\textit{Proprietary (API)}} \\
GPT-5.6-sol & 123.8 & 18.4 & 19.2 \\
GPT-5.4-mini & 120.0 & 31.1 & 31.7 \\
Claude Opus 4.8 & 296.9 & 21.2 & 22.0 \\
Claude Haiku 4.5 & 215.2 & 26.0 & 26.7 \\
Qwen3.6-Plus & 106.7 & 26.1 & 26.9 \\
Kimi-K2.6 & 172.8 & 22.8 & 23.6 \\
\midrule
\multicolumn{4}{l}{\textit{Open-weight}} \\
Qwen3.5-9B & 94.0 & 27.8 & 28.4 \\
Qwen3-VL-8B & 31.7 & 23.7 & 24.5 \\
Cosmos3-Nano-16B & 18.0 & 11.6 & 12.5 \\
RoboBrain2.5-8B & 42.4 & 32.1 & 32.6 \\
InternVL3.5-8B & 50.6 & 35.2 & 35.6 \\
\bottomrule
\end{tabular}
\end{table}

\section{Prompt Templates}

We provide the textual templates supplied to the MLLM planner. At every planning step, the textual input is paired with current egocentric visual observations. It identifies the controlled agent or team, states the grounded household goal, reports goal progress derived from live simulator state, enumerates the currently available action ids, and appends recent execution feedback and inventory state. The action menu is generated from scene metadata and task constraints, so the model selects an executable high-level skill rather than producing an unconstrained command.

The prompt encodes the main validity rules of the interface. An agent should locate an object before interaction, may pick up an object only when its hand is free and the object is accessible, must hold an object before placing or dropping it, and may open, close, toggle, or slice only compatible nearby objects. The planner should not repeat a successful \texttt{Find}, revise its choice after failed feedback, and return \texttt{DONE} only when live goal progress reports that every predicate is satisfied.

\subsection{Centralized Planner Prompt}

In the centralized setting, a single VLM planner observes all agents' egocentric views and chooses the next executable action for one agent at a time.

\begin{promptbox}{Centralized Planner Prompt}
\ttfamily
You are the centralized planner coordinating \{n\} agents (\{agents\}) in an AI2-THOR household task.\\
The image shows each agent's egocentric view side by side, left to right in the order: \{agents\}.\\
GOAL: \{goal\}

GOAL PROGRESS (live simulator state; trust this over your action history):\\
\{goal\_progress\}\\
\{coordination\_hint\_if\_any\}

Choose the single best NEXT action for ONE agent to make progress while coordinating\\
(divide work, respect ordering/dependencies, avoid two agents contending for the same spot/resource).\\
After Find succeeds, the agent is at the object; proceed to the interaction (PickUp/Put/Toggle/...),\\
do NOT repeat Find. Reply on two lines exactly:\\
REASON: \textless one short sentence\textgreater\\
ACTION: \textless the integer action id from the menu, or DONE if every goal is satisfied\textgreater

ACTION MENU (action\_id: skill [agent]):\\
\ \ \textless id\textgreater: agent\_i: find a Object [agent\_i]\\
\ \ \textless id\textgreater: agent\_i: explore Direction [agent\_i]\\
\ \ \textless id\textgreater: agent\_i: pick up the Object [agent\_i]\\
\ \ \textless id\textgreater: agent\_i: put object in hand on the Receptacle [agent\_i]\\
\ \ \textless id\textgreater: agent\_i: open / close the Receptacle [agent\_i]\\
\ \ \textless id\textgreater: agent\_i: turn on / turn off the ToggleableObject [agent\_i]\\
\ \ \textless id\textgreater: agent\_i: slice the SliceableObject [agent\_i]\\
\ \ ...

RECENT ACTIONS (most recent last):\\
\{recent\_action\_feedback\}

AGENT STATE:\\
\ \ agent\_1: holding=\{inventory\}\\
\ \ agent\_2: holding=\{inventory\}\\
\ \ ...
\end{promptbox}

The real prompt contains concrete action ids and object names, such as \texttt{47: agent\_1: pick up the CreditCard [agent\_1]}. The planner must return the integer action id rather than a free-form action string.

\subsection{Distributed Planner with Broadcast Prompt}

In the distributed setting, each agent is controlled by an independent VLM call and observes only its own egocentric view. Broadcast messages from previous rounds are included to support lightweight coordination.

\begin{promptbox}{Distributed Planner with Broadcast Prompt}
\ttfamily
You are \{me\}, one of \{n\} robots (\{agents\}) jointly doing an AI2-THOR household task.\\
You control ONLY yourself and you see ONLY your own egocentric view (the image above).\\
Your teammates are deciding their own next action at the same time; you will NOT see\\
their choice until next round, so coordinate by reasoning about what they are likely doing.

GOAL: \{goal\}

GOAL PROGRESS (live simulator state; trust this over your action history):\\
\{goal\_progress\}\\
\{coordination\_hint\_if\_any\}\\
TEAMMATE MESSAGES (most recent last):\\
\{broadcast\_messages\_from\_previous\_rounds\}

Choose the single best NEXT action for YOURSELF to make progress while coordinating\\
(divide the work, respect ordering/dependencies, do NOT contend with a teammate for the\\
same object / spot / resource). After Find succeeds you are at the object; proceed to the\\
interaction (PickUp/Put/Toggle/...), do NOT repeat Find. Reply on these lines exactly:\\
REASON: \textless one short sentence\textgreater\\
ACTION: \textless the integer action id from YOUR menu, or DONE if the whole task is satisfied\textgreater\\
MSG: \textless one short note telling teammates what you are doing / will do\textgreater

YOUR ACTION MENU (action\_id: skill):\\
\ \ \textless id\textgreater: agent\_i: find a Object\\
\ \ \textless id\textgreater: agent\_i: explore Direction\\
\ \ \textless id\textgreater: agent\_i: pick up the Object\\
\ \ \textless id\textgreater: agent\_i: put object in hand on the Receptacle\\
\ \ \textless id\textgreater: agent\_i: open / close the Receptacle\\
\ \ \textless id\textgreater: agent\_i: turn on / turn off the ToggleableObject\\
\ \ \textless id\textgreater: agent\_i: slice the SliceableObject\\
\ \ ...

YOUR RECENT ACTIONS (most recent last):\\
\{this\_agent\_recent\_feedback\}

YOUR STATE: holding=\{inventory\}
\end{promptbox}

\subsection{Execution Loop}

\begin{enumerate}
\item The environment renders visual observations and builds a compact action menu from scene metadata and task constraints.
\item The VLM chooses an action id. In C0 this is a global decision; in D1 each agent proposes one action concurrently.
\item The runner parses \texttt{ACTION: <id|DONE>} and resolves the id to a concrete skill call such as \texttt{PickUp(agent\_1, CreditCard)}.
\item AI2-THOR executes the skill, returning success/failure feedback.
\item The next prompt includes updated goal progress, recent action feedback, inventories, and, for D1, teammate broadcast messages from previous rounds.
\end{enumerate}

\section{Case Studies}

\begin{figure*}[t]
  \centering
  \includegraphics[width=\linewidth]{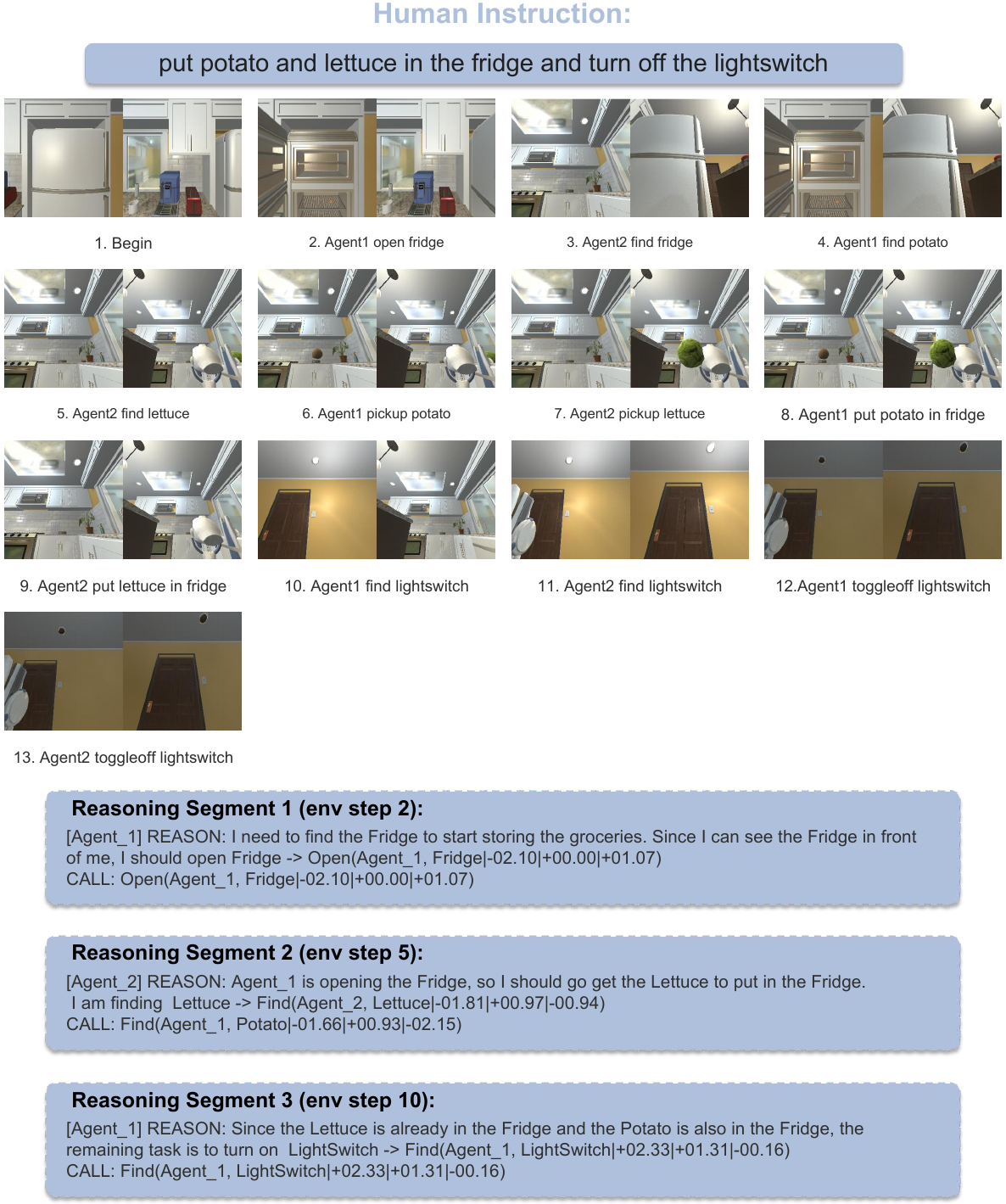}
  \caption{Successful planning for D1 task allocation. In \emph{Store Groceries}, agents divide the independent grocery-storage and light-switch subtasks, avoid duplicate work, and jointly satisfy all goal predicates.}
  \label{fig:case-task-allocation-store-groceries}
\end{figure*}

\begin{figure*}[t]
  \centering
  \includegraphics[width=\linewidth]{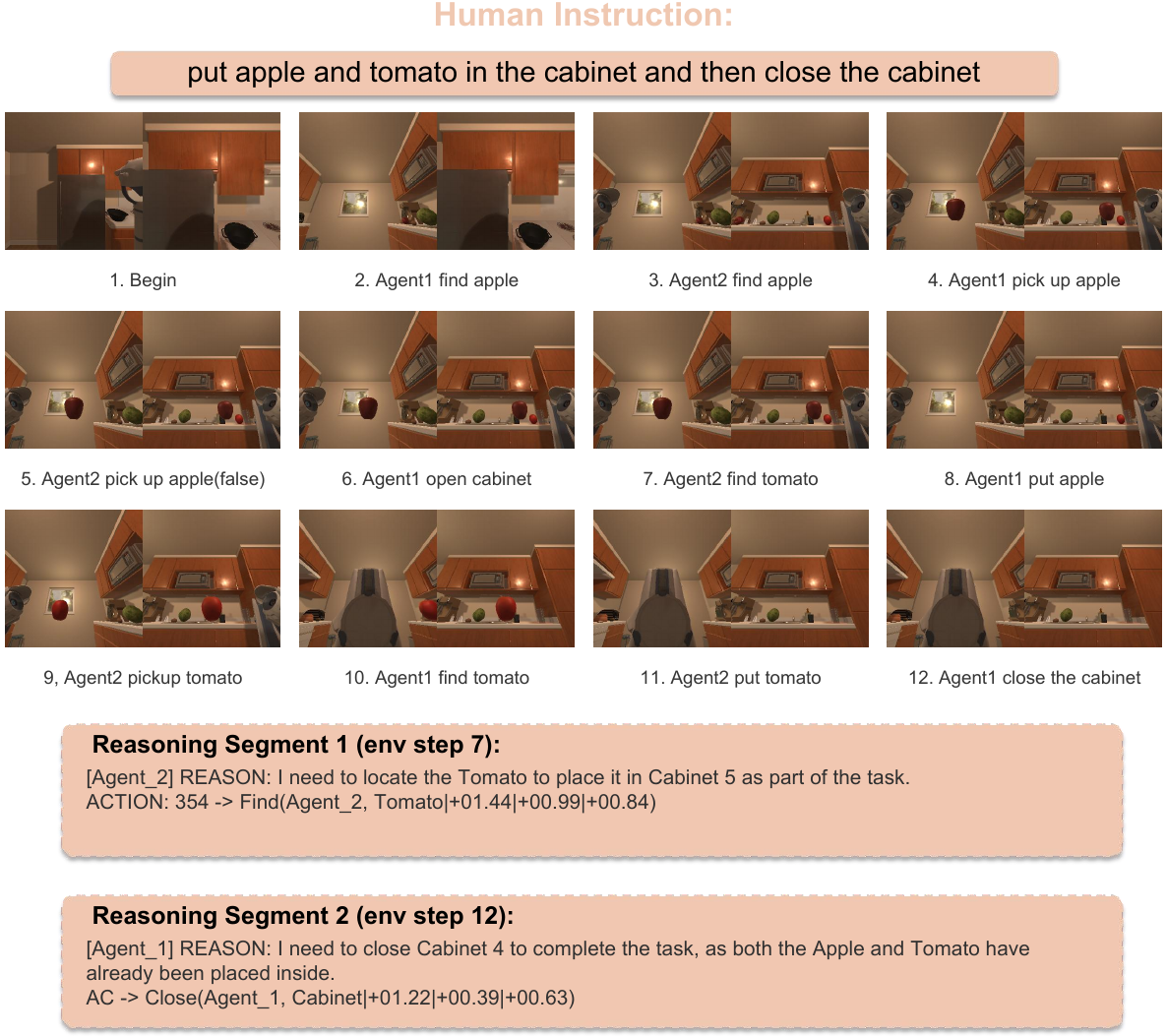}
  \caption{Successful planning for D2 sequential ordering. In \emph{Stock Cabinet}, the team preserves the open--load--close dependency and issues the terminal close action only after all required placements.}
  \label{fig:case-sequential-ordering-stock-cabinet}
\end{figure*}

\begin{figure*}[t]
  \centering
  \includegraphics[width=\linewidth]{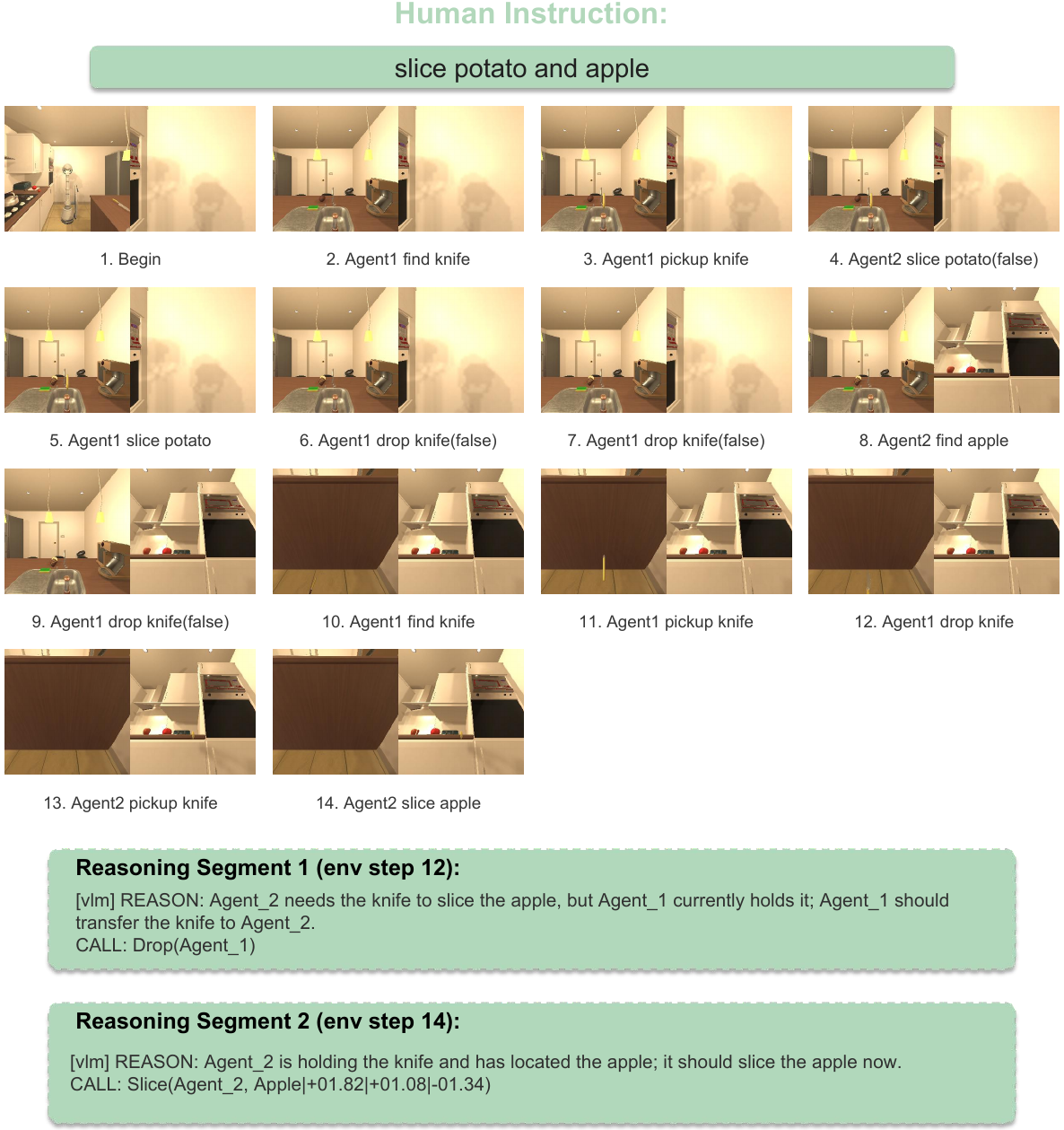}
  \caption{Successful planning for D3 mutual exclusion. In \emph{Prepare Food}, agents schedule access to the single shared Knife and complete their assigned slicing goals without resource collisions.}
  \label{fig:case-mutual-exclusion-prepare-food}
\end{figure*}

\begin{figure*}[t]
  \centering
  \includegraphics[width=\linewidth]{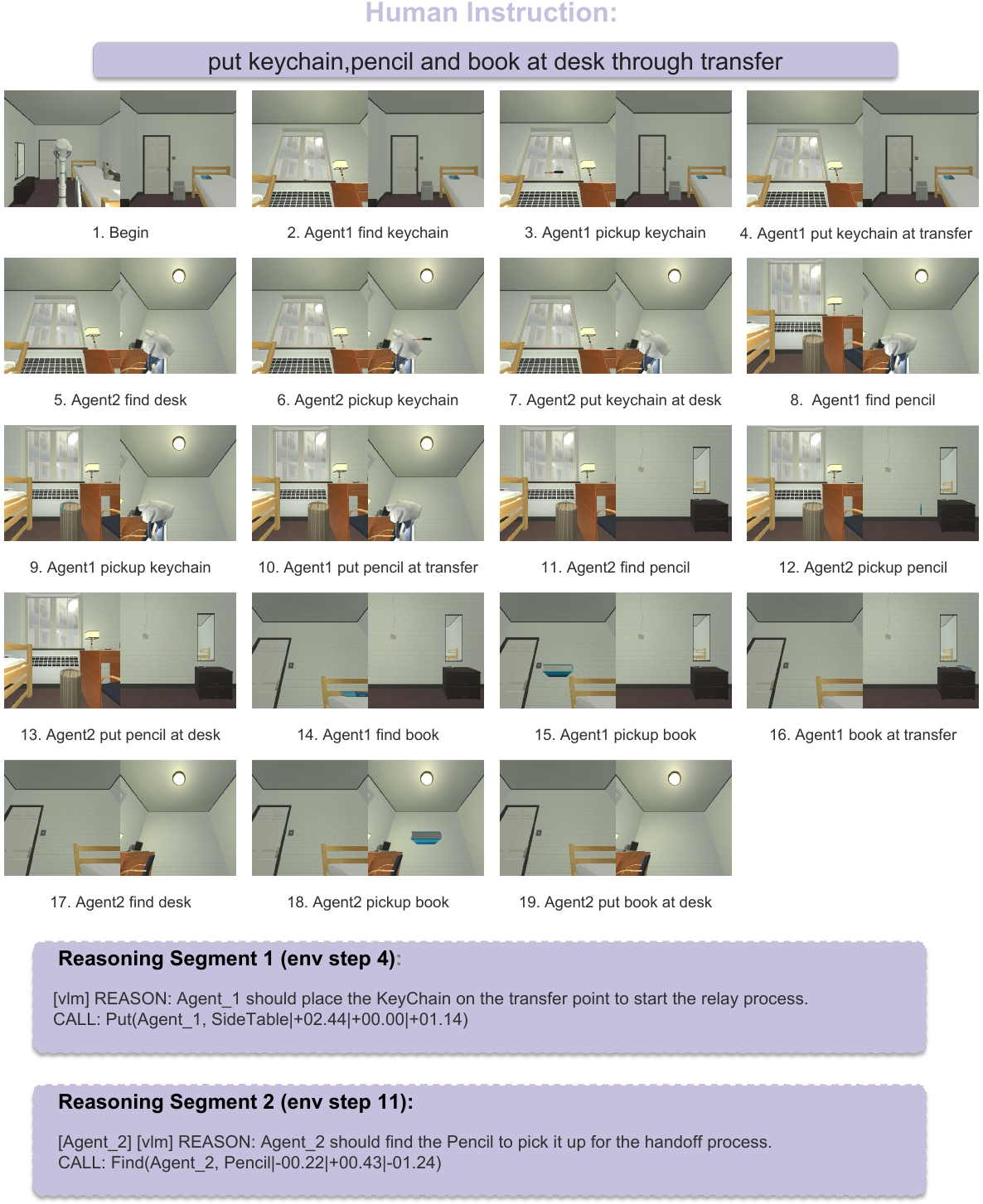}
  \caption{Successful planning for D4 handoff coordination. In \emph{Cross-Zone Handoff}, producer and consumer agents relay objects through an intermediate transfer point before completing final placement.}
  \label{fig:case-handoff-cross-zone-handoff}
\end{figure*}

We present four successful planning examples in Figures~\ref{fig:case-task-allocation-store-groceries}--\ref{fig:case-handoff-cross-zone-handoff}, with one trajectory for each coordination dimension. Each visualization aligns the agents' observations with their selected high-level actions and execution feedback, making it possible to inspect not only whether the final goal is reached but also how collaboration unfolds. Collectively, the cases show forms of coordination: distributing independent work, respecting a shared dependency, scheduling an exclusive tool, and maintaining a producer--consumer relay.

\paragraph{D1: Task Allocation.}
Figure~\ref{fig:case-task-allocation-store-groceries} illustrates a successful \emph{Store Groceries} trajectory. The conjunctive instruction contains separable food-placement and light-switch objectives. The agents commit to different pending subgoals and continue from the updated goal-progress state instead of redundantly pursuing an object already handled by a teammate. Their actions therefore overlap useful work and complete the household goal with a compact division of labor.

\paragraph{D2: Sequential Ordering.}
Figure~\ref{fig:case-sequential-ordering-stock-cabinet} shows how parallel work must be reconciled with a terminal dependency. Different agents can retrieve the required food items concurrently, but the Cabinet must remain open and available until every placement has succeeded. The trajectory delays the final \texttt{Close} action until the live progress state confirms that both items are inside, demonstrating shared precedence tracking rather than independent greedy execution.

\paragraph{D3: Mutual Exclusion.}
Figure~\ref{fig:case-mutual-exclusion-prepare-food} presents a shared-Knife food-preparation task. Although each agent has a distinct slicing target, the required tool can be held by only one agent at a time. The successful trajectory serializes Knife ownership while allowing other preparation actions to proceed, and the current holder yields access after completing its assigned slice. This avoids conflicting acquisition attempts and makes the exclusive resource available to every teammate that needs it.

\paragraph{D4: Handoff Coordination.}
Figure~\ref{fig:case-handoff-cross-zone-handoff} visualizes a multi-stage transport plan across separated zones. Source-side agents first deliver objects to the designated transfer location; destination-side agents then recognize the newly available buffer contents and carry them to the final receptacle. Repeating this producer--consumer pattern across objects prevents both premature downstream searches and excessive accumulation at the handoff point. The case highlights that D4 success depends on synchronized roles and intermediate state changes, not merely on independent point-to-point navigation.

\end{document}